%% file: main.tex
\documentclass[10pt,twocolumn,letterpaper]{article}

\usepackage{cvpr}               % To produce the CAMERA-READY version
\usepackage{graphicx}
\usepackage{amssymb}
\usepackage{amsfonts}
\usepackage{amsmath}
\usepackage{bm}
\usepackage{mathrsfs}
\usepackage{makecell}
\usepackage{multirow}
\usepackage{multicol}
\usepackage{booktabs}
\usepackage{cuted}
\usepackage{capt-of}
\usepackage{placeins}
\usepackage{stackengine} 
\usepackage{indentfirst}
\usepackage{algorithm}
\usepackage[noend]{algpseudocode}
\usepackage{tabularray}
\usepackage{setspace}
\usepackage[accsupp]{axessibility}

\RequirePackage{tabularray}

\usepackage{color}
\definecolor{darkgreen}{rgb}{0,0.5,0.7}
\definecolor{lightgreen}{rgb}{0.4,0.8,0.4}
\definecolor{lightorange}{rgb}{0.9,0.6,0.1}
\definecolor{darkpurple}{rgb}{0.5,0,0.7}
\definecolor{darkblue}{rgb}{0.5,0,0.7}
\definecolor{Comment}{rgb}{0.62,0.478,0.259}
\definecolor{DynamicNet}{rgb}{0.216,0.537,0.620}
\definecolor{RayRefinement}{rgb}{0.494,0.392,0.620}
\definecolor{Fthetal}{rgb}{0.573,0.224,0.196}
\definecolor{Sgtq}{rgb}{0.553,0.486,0.357}
\definecolor{St}{rgb}{0.749,0.565,0}
\newcommand{\draft}[2]{\textcolor{#1}{#2}}

\newcommand{\best}[1] {\textbf{\draft{red}{#1}}}
\newcommand{\second}[1] {\underline{\draft{blue}{#1}}}

\definecolor{jhred}{rgb}{1,0,0}

\algnewcommand\algorithmicforeach{\textbf{for each}}
\algdef{S}[FOR]{ForEach}[1]{\algorithmicforeach\ #1\ \algorithmicdo}


\definecolor{cvprblue}{rgb}{0.21,0.49,0.74}
\usepackage[pagebackref,breaklinks,colorlinks,allcolors=cvprblue]{hyperref}

\def\paperID{18280} % *** Enter the Paper ID here
\def\confName{CVPR}
\def\confYear{2025}

\title{MoDec-GS: Global-to-Local Motion Decomposition and Temporal Interval Adjustment for Compact Dynamic 3D Gaussian Splatting}

\author{\text{Sangwoon Kwak}$^{1,2}$,\quad \text{Joonsoo Kim}$^{1}$,\quad \text{Jun Young Jeong}$^{1}$,\quad \text{Won-Sik Cheong}$^{1}$, \\ 
\quad \text{Jihyong Oh}$^{3}$\footnotemark[2],\quad \text{Munchurl Kim}$^{2}$\footnotemark[2]\\
$^1$\text{Electronics and Telecommunications Research Institute,} \\
$^2$\text{Korea Advanced Institute of Science and Technology},\quad $^3$\text{Chung-Ang University} \\
{\tt\small $\{$s.kwak, joonsookim, jyj0120, wscheong$\}$@etri.re.kr}\\
{\tt\small $\{$sw.kwak, mkimee$\}$@kaist.ac.kr} \quad {\tt\small jihyongoh@cau.ac.kr}\\
{\tt\small \url{https://kaist-viclab.github.io/MoDecGS-site/}}
}

\begin{document}

\twocolumn[{
\maketitle
\begin{center}
\vspace{-0.5cm}
\includegraphics[scale=0.53]{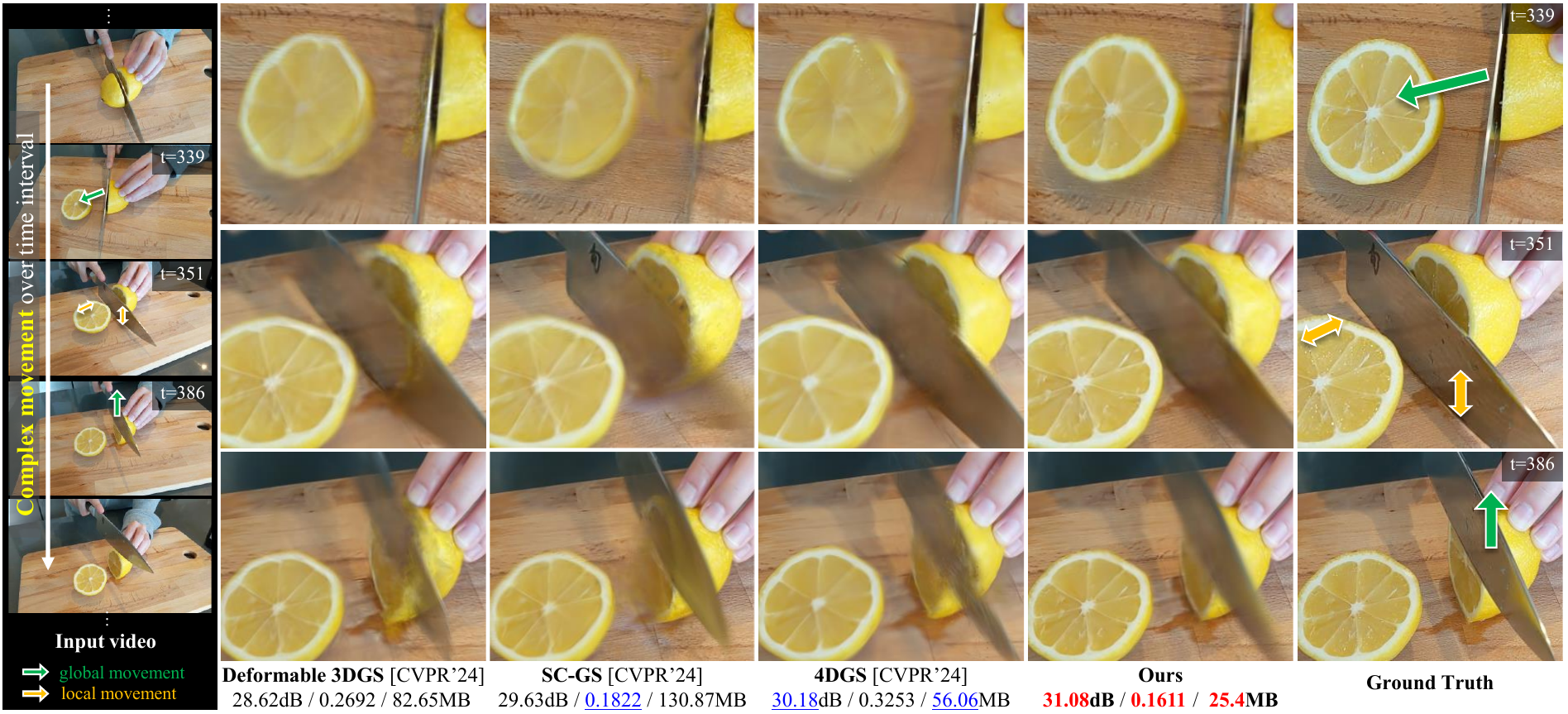}
\vspace{-0.15cm}
\captionof{figure}{\textbf{Novel view synthesis results on} \cite{Park2021HyperNeRF}. We introduce \textbf{MoDec-GS}, a novel framework for learning compact dynamic 3D Gaussians from real-world videos with complex motion. While existing SOTA methods \cite{Yang2024Deformable3DGS, Huang2024SCGS, Wu20244DGS} have difficulty modeling such complex combination of global and local motions, our approach effectively handles them thanks to GLMD (Sec. \ref{sect:method overview}), and outperforms the prior methods in rendering quality even with a \textit{compact model size}. The metrics under each framework are, PSNR (dB)$\uparrow$ / LPIPS \cite{zhang2018unreasonable} $\downarrow$ / Storage (MB)$\downarrow$.}
\label{fig:figure_page1}
\end{center}
}]

{
  \renewcommand{\thefootnote}%
    {\fnsymbol{footnote}}
  \footnotetext[2]{Co-corresponding authors.}
}

\input{sec/0_abstract}
\input{sec/1_intro}

\input{sec/2_related_works}
\input{sec/3_preliminary}

\input{sec/4_method}
\input{sec/5_experiment}
\input{sec/6_conclusion}

\vspace{2mm}

\begin{spacing}{0.8}
\noindent\textbf{Acknowledgements} 

\noindent{\footnotesize{This work was supported by the IITP grant funded by the Korea government (MSIT): Development of immersive video spatial computing technology for ultra-realistic metaverse services (No.2022-0-00022, RS-2022-II220022) and the Graduate School of Metaverse Convergence support program (RS-2024-00418847)}}
\end{spacing}
{
    \small
    \bibliographystyle{ieeenat_fullname}
    \bibliography{main}
}

\input{sec/X_suppl}


\end{document}

%% file: sec/0_abstract.tex
\vspace{-0.2cm}
\begin{abstract} 
3D Gaussian Splatting (3DGS) has made significant strides in scene representation and neural rendering, with intense efforts focused on adapting it for dynamic scenes. Despite delivering remarkable rendering quality and speed, existing methods struggle with storage demands and the representation of complex real-world motions. To address these challenges, we propose MoDec-GS, a memory-efficient Gaussian splatting framework designed to reconstruct novel views in challenging scenarios with complex motions. We introduce Global-to-Local Motion Decomposition (GLMD) to effectively capture dynamic motions in a coarse-to-fine manner. This approach leverages Global Canonical Scaffolds (Global CS) and Local Canonical Scaffolds (Local CS), which extend static Scaffold representation to dynamic video reconstruction. For Global CS, we propose Global Anchor Deformation (GAD) to efficiently represent global dynamics along complex motions, by directly deforming the implicit Scaffold attributes which are anchor position, offset, and local context features. Next, we finely adjust local motions via the Local Gaussian Deformation (LGD) of Local CS explicitly. Additionally, we introduce Temporal Interval Adjustment (TIA) to automatically control the temporal coverage of each Local CS during training, enabling MoDec-GS to find optimal interval assignments based on the specified number of temporal segments. Extensive evaluations demonstrate that MoDec-GS achieves an average 70$\%$ reduction in model size over state-of-the-art methods for dynamic 3D Gaussians from real-world dynamic videos while maintaining or even improving rendering quality.
%Our implementation will be publicly available upon publication. % \jh{[Our codes are available at: \url{https://anonymous.github.io/will-be-updated}]}
\end{abstract}

%% file: sec/1_intro.tex
\vspace{-0.4cm}
\section{Introduction}
Novel view synthesis (NVS) generates new perspectives of a scene from a limited set of images, closely approximating real footage. NVS has long been a key research area for many years, with advancements in techniques such as depth-image-based rendering \cite{Mori2009ViewGeneration, Fehn2004DIBR, Zinger2010FreeViewpoint}. This ongoing interest is largely driven by the broad applicability of NVS in areas such as virtual reality, augmented reality, and immersive media, where natural viewpoint transitions that mimic real-life experiences are essential for enhancing user realism \cite{broxton2020immersive, boyce2021mpeg}.

Approaching NVS as the task of modeling the radiance field has taken the computer vision community by storm. This paradigm shift, led by Neural Radiance Field (NeRF) \cite{Mildenhall2021Nerf}, has set a new photorealism standard that surpasses conventional methods. The original NeRF represents the radiance field as an implicit function linked to volume rendering, achieving remarkable visual fidelity. However, it faces challenges with its slow training and, more critically, rendering speed, which is far from real-time. Despite various optimization efforts \cite{Muller2022InstantNGP, keil2023Kplanes, liu2020neural}, achieving real-time rendering on consumer-level devices remains difficult, largely due to NeRF's reliance on pixel-wise volumetric rendering. 

Recently, 3D Gaussian Splatting (3DGS) \cite{kerbl20233dgs} has emerged as a compelling alternative, offering exceptional rendering speed without compromising visual quality. By representing the radiance field as a collection of 3D Gaussian ellipsoids, 3DGS enables efficient patch-wise rasterization through Gaussian projection and alpha compositing. This patch-level rasterization pipeline, fully leveraging GPU parallel computation, allows 3DGS to achieve unrivaled rendering speed. Subsequently, 3DGS has inspired diverse research trajectories, with extension to video inputs and compression emerging as key areas of focus \cite{li2024spacetime, yang2023real, Wu20244DGS, Lee2024Compact3DGS, lu2024scaffold, morgenstern2023compact}. 

% Since most real-world visual applications involve dynamic scenes, extending 3DGS to handle motion is a natural progression. Moreover, the explicit nature of 3DGS, comprising a large set of 3D Gaussian primitives with multi-dimensional attributes results in large data volumes, highlighting the need for compression.

Current approaches to dynamic adaptation often pair a static canonical 3DGS with implicit \cite{Wu20244DGS, duisterhof2023md, Liang2024GauFRe,  kratimenos2025dynmf, shaw2024swings, Huang2024SCGS, Yang2024Deformable3DGS} or explicit \cite{lin2024gaussian, katsumata2023efficient} components to manage the temporal deformation of the attributes within the canonical 3DGS. Another approach extends 3D Gaussian primitives into 4D by incorporating a temporal dimension \cite{duan20244d, yang2023real}. While both approaches preserve solid rendering quality for dynamic scenes, they face storage issues due to the multi-dimensional attributes assigned to numerous 3D Gaussians in canonical 3DGS or 4D Gaussians \cite{lee2024compact}. Handling long-duration content with complex motion also poses difficulties, as representing all frames with a unified model causes blurring due to its limited capacity, as noted by Shaw et al. \cite{shaw2024swings}. To address this, they segment sequences based on scene motion and train a separate model for each segment. However, this approach requires an extra step to compute motion vectors, which diminishes the usability of 3DGS. 

Early methods to address the storage demands of 3DGS had focused on compressing the original 3DGS representation, employing techniques like vector quantization \cite{navaneet2023compact3d,  fan2023lightgaussian, Lee2024Compact3DGS, niedermayr2024compressed}, Gaussian pruning \cite{fan2023lightgaussian, Lee2024Compact3DGS}, and implicit encoding of Gaussians' attributes \cite{Lee2024Compact3DGS, girish2023eagles, wu2024implicit}. Some recent studies have successfully introduced more memory-efficient representations based on the 3DGS framework, with a notable example being the anchor-based representation \cite{lu2024scaffold, wang2024contextgs, chen2025hac, liu2024compgs}, which assigns implicit features at sparse anchor points to predict attributes for a broader set of neighboring 3D Gaussians. However, extending these methods, originally designed for static scenes, to dynamic videos may be challenging, as dynamic scene modeling typically requires substantial additional components. A few recent methods have aimed to unify dynamic extension and memory efficiency, but one \cite{sun20243dgstream} is limited to multi-view sequences, while another \cite{lee2024compact} still struggles with long-duration content.

To address the limitations of existing methods, we propose a novel dynamic 3D Gaussian splatting framework, enhancing model compactness and rendering quality while preserving real-time capability. Our framework employs a deformation-based approach with an anchor-based representation \cite{lu2024scaffold} for the canonical 3DGS due to its compactness. Building on this, we introduce a two-stage deformation process inspired by the observation that natural motion involves both global and local components. In the first stage, called Global Anchor Deformation (GAD), the canonical representation is deformed to a specific time interval using an anchor deformation encoder that captures global motion across the entire sequence. In the second stage, called Local Gaussian Deformation (LGD), the deformed representation is refined via local deformation encoder capturing finer motion within the specific time interval around the chosen time point. A final feature of our framework is Temporal Interval Adjustment (TIA), which assigns the temporal interval to each deformation encoder and is automatically determined during training. This does not require any precomputed external information such as optical flow, and effectively localizes the dynamic motion of the scene. In short, our contributions are as follows: 
\begin{itemize}
    \item We propose a novel framework MoDec-GS based on Global-to-Local Motion Decomposition (GLMD), which effectively handles real world's complex motions composed of global and local movements.
    \item We introduce the TIA to adaptively control the temporal intervals of each local canonical anchor during training, enabling our MoDec-GS to achieve optimal visual quality even with a compactly limited model size. 
    \item Extensive experiments on three widely used monocular datasets show that our method significantly reduces storage while maintaining or even improving visual quality, specifically, on iPhone dataset \cite{Gao2022Dycheck}, it shows a PSNR gain of + 0.7dB and a storage reduction of -94$\%$ compared to the second best method in terms of quality, SC-GS \cite{Huang2024SCGS}.
\end{itemize}
  
\vspace{-0.1cm}

%% file: sec/2_related_works.tex
\section{Related Works}
\label{sec:formatting}

%-------------------------------------------------------------------------
\subsection{3D Gaussian Splatting for Dynamic Scenes}
A natural evolution of 3DGS \cite{kerbl20233dgs} for static scenes is its extension to dynamic scenes, with recent research in this area generally split into two main categories: deformation-based and 4D Gaussian-based methods. Deformation-based methods rely on a static canonical 3DGS, paired with a component that captures the temporal deformation of the attributes within this canonical representation. This deformation can be modeled implicitly using structures like MLP \cite{Liang2024GauFRe, Huang2024SCGS, Yang2024Deformable3DGS} or feature grids \cite{Wu20244DGS, duisterhof2023md}, or explicitly via polynomial \cite{li2024spacetime, lin2024gaussian}, Fourier \cite{lin2024gaussian, katsumata2023efficient} or learned basis functions \cite{kratimenos2025dynmf}. In contrast, 4D Gaussian-based methods \cite{yang2023real, duan20244d} introduce time as an extra dimension in the 3D Gaussian formulation.  Both works aim to integrate the 4D Gaussian paradigm within the established 3DGS training and rendering framework, with a particular focus on efficiently representing rotations in 4D space. 

Although both categories achieve decent rendering quality for dynamic scenes, they still demand substantial storage, to handle the multi-dimensional attributes of millions of 3D Gaussians in canonical 3DGS or 4D Gaussians \cite{zhang2024mega, lee2024compact}. Additionally, representing long-duration or large-motion contents with a compact model is challenging \cite{shaw2024swings}. This difficulty arises from the need to maintain fast rendering speed while ensuring reliable training on a limited set of sampled videos. A straightforward approach is to train a separate model for each empirically determined interval; however, determining the optimal interval empirically is inefficient, as the flow of motion information can vary significantly depending on the content. Shaw et al. \cite{shaw2024swings} proposed a method for temporal segmentation to address this issue, but this considers only the magnitude of motion. In contrast, we integrates a temporal segmentation method directly into the training process, enabling MoDec-GS to explore optimal intervals by itself to achieve the best rendering quality. 

\subsection{Compact 3D Gaussian Splatting}
To address the substantial memory demands of 3DGS, various strategies have been proposed. The first category focuses on compressing the original 3DGS representation, with key approaches including vector quantization \cite{navaneet2023compact3d,  fan2023lightgaussian, Lee2024Compact3DGS, niedermayr2024compressed}, pruning redundant Gaussians \cite{fan2023lightgaussian, Lee2024Compact3DGS, wang2024end}, implicit encoding of high-dimensional attributes \cite{Lee2024Compact3DGS, girish2023eagles, wu2024implicit}, using standardized compression pipelines \cite{morgenstern2023compact, wu2024implicit, fan2023lightgaussian} and applying entropy constraint \cite{wang2024end}. The second category explores more efficient Gaussian representations to mitigate storage challenges \cite{hamdi2024ges, lu2024scaffold}. A prominent example is Scaffold-GS, which introduces a unique method by assigning learnable features to a sparse set of anchor points that predict attributes for a broader set of neighboring 3D Gaussians. Recent advancements in the Scaffold-GS framework have further enhanced memory efficiency by organizing anchor points hierarchically across multiple levels \cite{wang2024contextgs} or using a binary hash grid to model context for unstructured anchor attributes \cite{chen2025hac}. 

However, adapting methods from both categories to 4DGS may not be straightforward, as most 4D extensions require substantial architectural modifications to extend 3DGS for dynamic scene modeling. A few recent approaches have covered both dynamic extension and memory efficiency; for instance, Sun et al. \cite{sun20243dgstream} propose a framework for on-the-fly training, where adaptive control over the quantity of 3D Gaussians is employed, allowing the model size to remain moderate for streaming. However, this method assumes multi-view inputs. Lee et al. \cite{lee2024compact} implement a combination of compression techniques, including residual vector quantization and hash grid-based encoding, on top of the Spacetime Gaussian proposed by Li et al. \cite{li2024spacetime}. However, this approach doesn't provide an efficient solution for handling long-duration, large-motion content. 

%% file: sec/3_preliminary.tex
\vspace{-0.1cm}
\section{Preliminary}

\begin{figure*}[t!]
\centering
\includegraphics[scale=0.52]{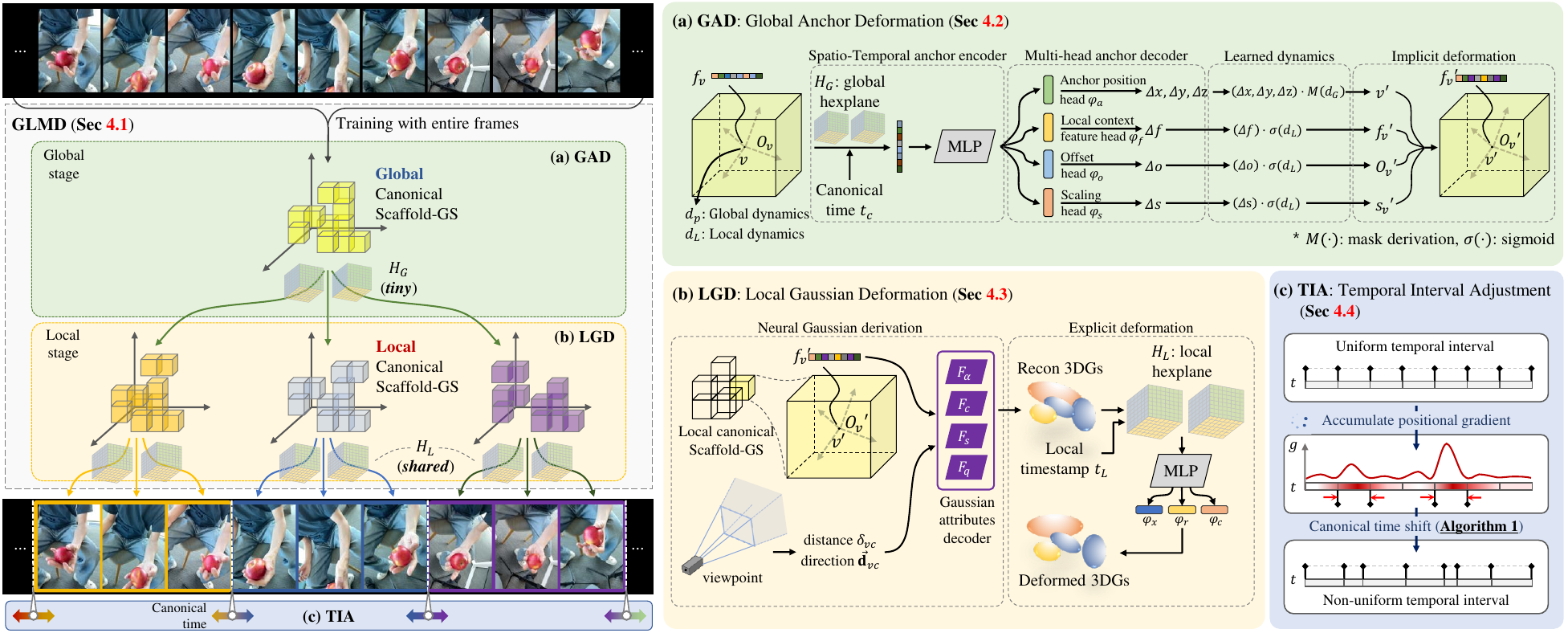}
\vspace{-0.3cm}
\caption{\textbf{Overview of our MoDec-GS framework.} To effectively train dynamic 3D Gaussians with complex motion, we introduce Global-to-Local Motion Decomposition (GLMD) (Sec \ref{sect:method overview}). We first train a Global Canonical Scaffold-GS (Global CS) with entire frames, and apply a Global Anchor Deformation (GAD) to Local Canonical Scaffold-GS (Local CS) dedicated to represent its corresponding temporal segment (Sec \ref{sect4.2}). Next, to finely adjust the remaining local motion, we apply Local Gaussian Deformation (LGD) which explicitly deforms the reconstructed 3D Gaussians with a shared hexplane (Sec \ref{sect4.3}). During the training, Temporal Interval Adjustment (TIA) is performed, optimizing the temporal interval into a non-uniform interval that adopts to the scene's level of motion (Sec \ref{sect4.4}).}
\label{fig:Method_overview}
\vspace{-2mm}
\end{figure*}
%%%%%%%%%%%%%%%%%%%%%%%%%%%%%%%%%%%%%%%

%%%%%%%%%%%%%%%%%%%%%%%%%%%%%%%%%%%%%%%%%%%%%
\subsection{Splatting of Gaussian primitives}
Gaussian primitives, or Gaussians, are characterized using their respective rotation matrices $R$ and scaling matrices $S=diag([s_{i}])$. To render the primitives for a viewport that corresponds to a viewing transformation matrix $W$, one can calculate the following covariance matrix:
\vspace{-1mm}
\begin{equation} \label{eq:2D conic matrix}
    \Sigma^{\prime} = J^{T} W^{T} \Sigma W J,
\end{equation}
\noindent where $\Sigma = R^{T}S^{T}SR$ and $J$ is an affine approximation of the projective transformation \cite{kerbl20233dgs}. The covariance matrix represents the approximate shape of the projected Gaussian. Once the 2D covariance matrices and the projected central positions for each Gaussians are calculated, the primitives are sorted in the order of the depth values. Lastly, the colors for each pixel in the viewport can be calculated by the following alpha-blending procedure: 
\vspace{-1mm}
\begin{equation} \label{eq:alpha_blending}
     \bm{C} = \sum^n_{i=1}\bm{c}_i\alpha_i\prod^{i-1}_{j=1}(1-\alpha_j),
\end{equation}
\noindent where $\mathbf{c}_{i}$ is the color of the $i$-th primitive that is calculated from spherical harmonic coefficients. The opacity $\alpha_{i}$ is obtained by evaluating the 2D Gaussian distribution function at the pixel position.

\subsection{Scaffold-GS}
Scaffold-GS representation consists of a set of anchor points which are the central points of voxels of a predefined size, a set of neural Gaussians associated with the anchor points, and a set of neural networks that predict the attributes of the neural Gaussians. There are $k$-number of neural Gaussians that are associated with each anchor point, and such a group of Gaussians that are softly bound to a spatial point work as a local representation. The centers of the neural Gaussians are given as follows:
\vspace{-1mm}
\begin{equation} \label{eq:scaffold_neural_gaussian_centers}
    \mathbf{m}_{i} = \mathbf{x}_{v} + \mathbf{o}_{i},
\end{equation}
\noindent where $i \in \{0, ..., k-1\}$ and $\mathbf{x}_{v}$ is the position of an anchor point $v$. $\mathbf{m}_{i}$ and $\mathbf{o}_{i}$ are the center position and the learnable offset for the $i$-th neural Gaussian. The opacities of the neural Gaussians are predicted as follows:
\vspace{-1mm}
\begin{equation} \label{eq:scaffold_neural_gaussian_opacity}
    \{ \text{attr}_{v,0}, \cdots, \text{attr}_{v,k-1}\} = F_{\text{attr}}(\hat{f}_v, \delta_{v, \text{cam}}, \overrightarrow {\textbf{d}}_{v, \text{cam}})
\end{equation}
\noindent where $\text{attr}_{v,i}$ is the attribute of the $i$-th neural Gaussian associated with the anchor point $v$. The attributes include opacity, color, quaternion, and scale. Separate neural networks $F_{\text{attr}}$ are used to predict the attributes where the networks take inputs including learnable anchor feature $\hat{f}_v$ and the displacement $(\delta_{v, \text{cam}}, \overrightarrow {\textbf{d}}_{v, \text{cam}} )$ from the viewing position to the anchor $v$. Similar to the densification in 3DGS, the anchor points are added or removed based on the gradient accumulation and the opacity. Please see \textit{Suppl.} \ref{sub:anchor-based representation} for detail.

%% file: sec/4_method.tex
% Abstract에서 정정한 용어들로 모두 싹 통일
% 전개 흐름도 Abstract하고 맞춰서
% 그림 2 수정
% 전체적으로 한번만 가다듬고, Algorithm 1 채우고, Experiment로 바로 넘어가자
\vspace{-1mm}
\section{Proposed Method}
%---------------------------- overview --------------------------%
\vspace{-1mm}
\subsection{Overview of MoDec-GS}
\vspace{-1mm}
\label{sect:method overview}
% Our goal is to develop a dynamic 3D Gaussian splatting framework for complex motion-contained dynamic videos that specifically addresses rendering quality and \textit{compact} model size without sacrificing real-time rendering capabilities. To this end, w
We adopt a deformation-based real-time dynamic scene rendering method \cite{Wu20244DGS}, but use anchor-based representation \cite{lu2024scaffold} as a canonical 3DGS due to its compactness. To effectively capture real-world videos with a complex combination of global and local motions, we introduce Global-to-Local Motion Decomposition (GLMD) as illustrated in Fig. \ref{fig:Method_overview}. GLMD consists of two stages: the first models global motion, while the second refines local motion. In the first stage, we apply Global Anchor Deformation (GAD), which deforms the position and attributes of anchors with a tiny global hexplane, transforming the Global Canonical Scaffold-GS (Global CS) into the Local Canonical Scaffold-GS (Local CS) (Sec. \ref{sect4.2}). As shown in Fig. \ref{fig:2stage_deformation}, this anchor-based transformation effectively captures global motion. Additionally, we embed learnable parameters into anchors to reflect motion characteristics, enabling effective control over both anchor-wise global motion and local motion within each anchor. In the second stage, the Local CS is reconstructed into 3DGs through neural Gaussian derivation and then explicitly deformed to each target timestamp using a shared local hexplane (Sec.  \ref{sect4.3}). To optimize the temporal interval assigned to each Local CS based on the scene motion, we propose Temporal Interval Adjustment (TIA) (Sec.  \ref{sect4.4}). This method dynamically re-balances temporal intervals during training, efficiently utilizing limited representation capability, without requiring any precomputed external information such as optical flow or tracking \cite{liu2024swings, liu2024modgs, lei2024mosca}.

\subsection{Global Anchor Deformation (GAD)}
\begin{figure}
\centering
\includegraphics[scale=0.5]{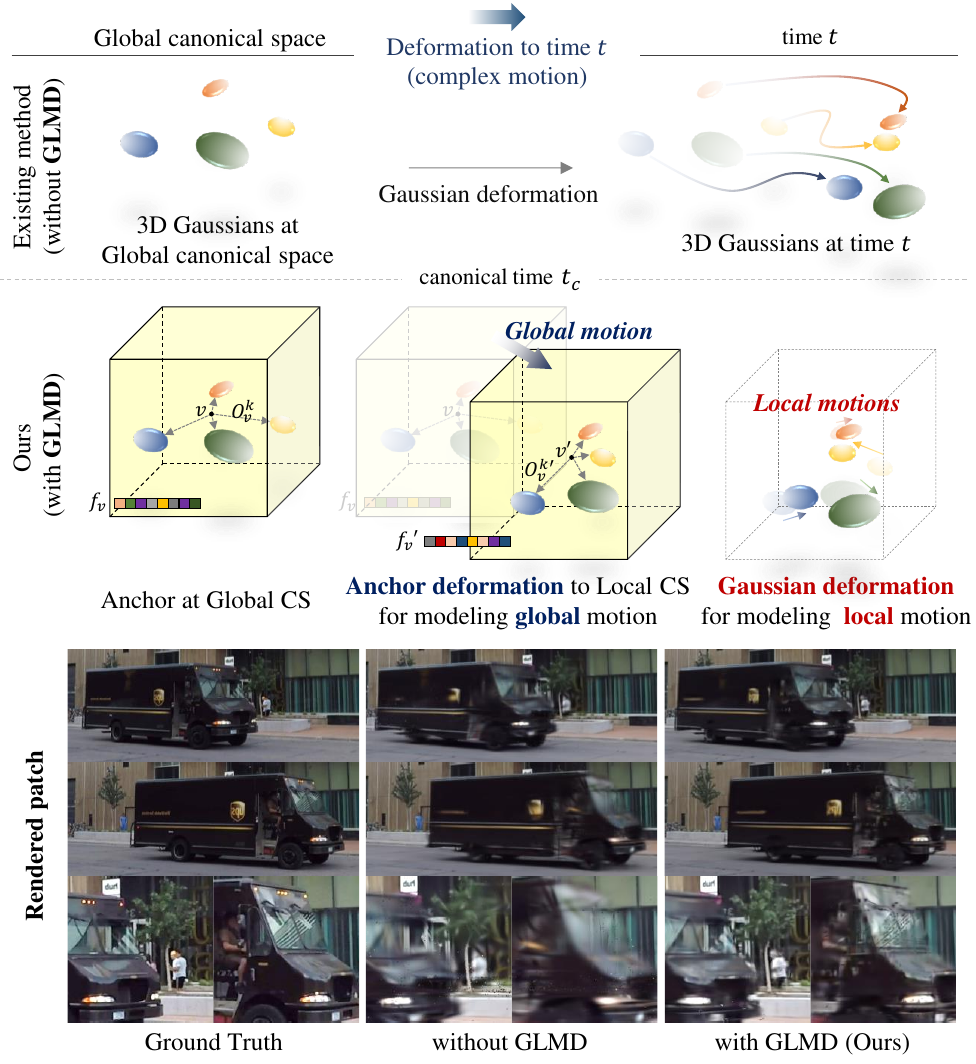}
\caption{\textbf{Concept and effect of 2-stage deformation.} For representing a complex motion of 3D Gaussians, a global movement over time intervals can be more efficiently handled through deformation of anchor itself. In contrast, subtle motions of individual 3D Gaussians within a time interval can be effectively addressed by explicit deformation of each Gaussian.}
\label{fig:2stage_deformation}
\vspace{-0.5cm}
\end{figure}

%---------------------------- Method 4.2 --------------------------%
\label{sect4.2}
% '두가지 있다'는 흐름보다는 기존에는 '(i)' 만 있었는데 이게 complex combination motion에는 잘 안될때도 있다. --> 이걸 두 단계로 나눠서 하면 좀 더 잘 모델링 된다. 방법론은 Fig. 3  성능은 See ection ** 보면 된다)
\noindent {\textbf{Anchor Deformation}.}\quad One approach to representing dynamic motion based on deformation is learning a hexplane that deforms 3DGs attributes after reconstruction \cite{Wu20244DGS}. While intuitive and efficient, it may struggle to handle a complex combinations of global and local motions due to the hexplane's limited capacity. In contrast, another method to achieve this involves directly deforming the anchor's position and attributes in anchor-based representation \cite{lu2024scaffold}. As shown in Fig \ref{fig:2stage_deformation},  the method of deforming the anchor itself is more efficient for representing a global motion of relatively large objects, rather than learning deformation fields for each reconstructed individual 3DGs. Therefore, as shown in Fig. \ref{fig:Method_overview}-(a), we deform the anchor's position and its attributes in GAD stage. For a given anchor point $v$, the anchor position $x_v,y_v,z_v$ is queried in a tiny global hexplane $H_G$ along with a timestamp. Here, the timestamp corresponds to the canonical time $t_c$, which is a time representing each divided temporal segment, determining temporal interval of the Local CS. The queried feature is decoded by a tiny MLP and a multi-head anchor decoder, producing the deformations for the position and attributes associated with the anchor: $(\Delta x, \Delta y, \Delta z)$, $\Delta f_v$, $\Delta O_v$, $\Delta s_v$. For example, a deformation of local context feature can be obtained by
\vspace{-1mm}
\begin{equation} \label{eq:dynamics masking}
   \Delta f_v = \varphi_f[F_G(H_G(x_v,y_v,z_v, t_c)],
\end{equation}
\noindent where $F_G$ is a light MLP and $\varphi_f$ is a local context feature head among the multi-head anchor decoders. The deformation values are added to the anchor attributes, producing a deformed anchor, at which point the proposed novel term, learnable \textit{anchor dynamics}, are incorporated. 

\noindent \textbf{Anchor Dynamics.}\quad Even for a long-range video, a considerable portion of the scene is still static. Rather than separately generating static and dynamic parts \cite{Liang2024GauFRe} or utilizing a precomputed external dynamic mask \cite{luiten2024dynamic}, we aim to learn motion dynamics by assigning additional learnable attributes to the anchor, allowing them to be optimized during the training process. To separately model the global movement characteristics of the anchor and the local movements within the anchor, we applied and trained $d_G$ and $d_L$ independently. Global dynamics $d_G$ learns whether the entire anchor moves globally and applies binary masking based on a threshold. This learnable masking inspired by \cite{Lee2024Compact3DGS} is derived as follows: 
\vspace{-1mm}
\begin{equation} \label{eq:dynamics masking}
    M(d_G)=\text{sg}(\mathcal{I}[\sigma(d_G)>\epsilon]-\sigma(d_G)) + \sigma(d_G),
\end{equation}
\noindent where $\text{sg}(\cdot)$ is the stop gradient operator, $\mathcal{I}$ is an indicator, $\sigma(\cdot)$ is sigmoid function, and $\epsilon$ is the masking threshold. Local dynamics $d_L$ is simply activated and then multiplied to the attributes of the corresponding anchor. Finally, the attributes of the deformed anchor are given by
\vspace{-1mm}
\begin{align} \label{eq:dynamics masking}
    x_{v'},y_{v'},z_{v'} = (x_v, y_v, z_v)& + M(d_G)\cdot(\Delta x,\Delta y,\Delta z) \\
    f_{v'} = f_v &+ \Delta f \cdot \sigma(d_L), \\
    o_{v'} = o_v &+ \Delta o \cdot \sigma(d_L), \\
    s_{v'} = s_v &+ \Delta s \cdot \sigma(d_L). 
\end{align}
\vspace{-0.3cm}
\vspace{-0.3cm}
\subsection{Local Gaussian Deformation (LGD)}
%---------------------------- Method 4.3 --------------------------%
\label{sect4.3}
%After global movements of each anchor over time intervals are well-captured in the first stage, the remaining movements of individual 3D Gaussians are expected to become relatively small and simplified, 
Once global motion over a time interval is captured in the first stage, the remaining local motion of individual 3D Gaussians are relatively minor and simplified, as shown in Fig. \ref{fig:2stage_deformation}. This claim is also supported by \textit{Suppl.} \ref{sup:visualization of GLMD}. Representing such movements can be effectively handled by the explicit deformation \cite{Wu20244DGS} of the reconstructed Gaussians, rather than anchor deformation, 
since it would require learning feature changes capable of generating the attributes of the displaced Gaussians (See Sec. \ref{table:ablation_study} and Tab. \ref{table:ablation_study}). Based on this reasoning, a deformed Local CS is first reconstructed into 3D Gaussians through neural Gaussian derivation. 

\noindent \textbf{Neural Gaussian Derivation.}\quad Within a given view frustum, $k$ neural Gaussians are spawned from the deformed anchor, and each Gaussian's attributes are reconstructed using the deformed feature along with the viewing direction and distance. For instance, an opacity set of $k$ Gaussians is spawned as follows:
\vspace{-1mm}
\begin{equation} \label{eq:method_neural_gaussian_derivation}
  \{ \alpha_0, \cdots,\alpha_{k-1} \}=F_{\alpha}(\hat{f'}_v, \delta_{v', \text{cam}}, \overrightarrow {\textbf{d}}_{v', \text{cam}}),
\end{equation}
\noindent where $F_\alpha$ is MLP decoder, $\hat{f'}_v$ is a feature bank constructed from the deformed feature on anchor $v'$, $\delta_{v', \text{cam}}$ and $\overrightarrow {\textbf{d}}_{v', \text{cam}}$ are relative distance and viewing direction from viewpoint to the anchor, respectively \cite{lu2024scaffold}.  

\noindent \textbf{Gaussian Deformation.}\quad The spawned neural Gaussians are then explicitly deformed to the target timestamp \cite{Wu20244DGS}. For example, positional deformations of $k$-th neural Gaussian in a Local CS is given by:
\vspace{-1mm}
\begin{equation} \label{eq:dynamics masking}
   \Delta x_k, \Delta y_k, \Delta z_k = \varphi_p[F_L(H_L(x_k,y_k,z_k, t_L)]
\end{equation}
\noindent where $H_L$ is a local multi-resolution hexplane, $F_L$ is a MLP decoder, $\varphi_p$ is a position head, and $t_L$ is a target timestamp. Note that the local hexplane and corresponding MLP decoder are shared across each Local CS for compactness. 

\subsection{Temporal Interval Adjustment (TIA)}
%---------------------------- Method 4.4 --------------------------%
\label{sect4.4}
To adaptively localize the degree of motion and guide the temporal scope of each Local CS, we divided the total frames $N$ into $l$ segments, where $1 < l < N$. Initially, segments have uniform temporal intervals. However, depending on the scene motion characteristics, their optimal sizes that each Local CS can effectively represent may vary. To account for this, we propose Temporal Interval Adjustment (TIA) to adjust the temporal intervals to fit the scene during the training process. The TIA re-balances the complexity of deformation required for each local canonical Gaussians, enabling effective scene representation even with \textit{compactly} limited size of hexplane. 

\noindent \textbf{Canonical time shift.} For the temporal interval adjustment, we employ the canonical time-based shift method, as illustrated in Fig. \ref{fig:Method_overview}-(c). 
Basically, temporal intervals are managed by a canonical time list $T_c = [t_1, t_2, \cdots, t_{l-1}]$, which represents the lowest timestamp of each interval and serves as the boundary between temporal segments. During the training process, this list is fixed at equal intervals from the normalized entire time range $[0,1]$, until a preset iteration for starting temporal adjustment, $T^{\text{TIA}}_{\text{from}}$. After the starting iteration, the TIA process, as described in Algo. \ref{alg:Temporal_Interval_Adjustment}, is repeated until a preset iteration for ending the process, $T^{\text{TIA}}_{\text{until}}$. 
During each adjustment period $T^{\text{TIA}}_{\text{period}}$, positional gradients are accumulated in the temporal interval to where the timestamp of each training view belongs. The accumulated gradient list $\textbf{G}^{\text{acc}} = [g^{\text{acc}}_{0},g^{\text{acc}}_{1}, \cdots, g^{\text{acc}}_{l-1}]$ and the accumulation count list $\nu^{\text{acc}} = [\nu^{\text{acc}}_{0},\nu^{\text{acc}}_{1}, \cdots, \nu^{\text{acc}}_{l-1}]$, are initially set to zero, and updated in each iteration as follows: 
\vspace{-1mm}
\begin{align} \label{eq:gradient accumulation}
   g^{\text{acc}}_{c} &= g^{\text{acc}}_{c} + g^{pos}_t, \\
   \nu^{\text{acc}}_{c} &= \nu^{\text{acc}}_{c} + 1,
\end{align}
\noindent where $g^{pos}_t$ is the Frobenius norm of positional gradient for a certain iteration where the training view has timestamp $t$ within $[t_{c}, t_{c+1})$. Please note that the left boundary of the first temporal interval should not be represented as an adjustable canonical time but always be fixed at $0$. Therefore, $T_c$ is one element shorter in length than $\textbf{G}^{\text{acc}}$. Based on the statistics of the accumulated gradients, we identify Local CS with insufficient expressiveness, and shrink the corresponding temporal intervals. The idea behind this approach is that temporal segments with significantly high accumulated positional gradient indicate regions where the Local CS struggles to represent efficiently; therefore, we reduce their assigned time intervals. 
Each temporal interval with an accumulated gradient greater than the preset threshold $\tau_{\text{TIA}}$ is shrunk by a step size $s_{\text{TIA}}$ on both sides. 
\vspace{-2mm}
%--------------------------------Algorithm start ------------------------------------%
\begin{algorithm}
\small
\caption{Temporal Interval Adjustment (Fig. \ref{fig:Method_overview}-(c))} 
\begin{algorithmic}[1]
\Procedure{TIA}{$T_c, \mathbf{G}^{\text{acc}}, \nu^{\text{acc}}, g^{pos}_t, \tau_{\text{TIA}}, s_{\text{TIA}}$}
    \If {$T^{\text{TIA}}_{\text{from}}$ $\leq$ iter $\leq$ $T^{\text{TIA}}_{\text{until}}$} 
        \State Update $\mathbf{G}^{\text{acc}}$ with $g^{pos}_t$ (Eq. \ref{eq:gradient accumulation})
        \If {iter $\%$ $T^{\text{TIA}}_{\text{period}}$ $= 0$}
            \State $\mu = \sum_{c=0}^{l-1} (g^{\text{acc}}_{c} / \nu^{\text{acc}}_{c}) / l $ \Comment{\textcolor{Comment}{acc. grad. mean}}
            \State $\sigma = \sqrt{\sum_{c=0}^{l-1}[(g^{\text{acc}}_{c} / \nu^{\text{acc}}_{c})-\mu]^2 / l}$ \Comment{\textcolor{Comment}{std.}}
            \For {$j=0$ to $l-1$}
                \If{$g^{\text{acc}}_{j} \geq \mu + \tau_{\text{TIA}}\cdot\sigma$} \Comment{\textcolor{Comment}{shrink}}
                    \If{$j \neq 0$ \textbf{and} $t_{j} \leq t_{j+1} - s_{\text{TIA}}$}
                        \State $t_{j} \leftarrow t_{j} + s_{\text{TIA}}$ 
                    \EndIf
                    \If{$j \neq l-1$ \textbf{and} $t_{j} \leq t_{j+1} - s_{\text{TIA}}$}
                        \State $t_{j+1} \leftarrow t_{j+1} - s_{\text{TIA}}$
                    \EndIf
                \EndIf
            \EndFor
        \State \textbf{Init} $\mathbf{G}^{\text{acc}}$, $\nu^{\text{acc}}$, $\mu$, $\sigma$
        \EndIf
    \EndIf
\EndProcedure
\end{algorithmic}
\label{alg:Temporal_Interval_Adjustment}
\end{algorithm}
\vspace{-2mm}
%--------------------------------Algorithm end ------------------------------------%

%% file: sec/5_experiment.tex
% 구성 흐름은, dataset, metric, comparison methods 간략 설명 
% --> Main results (Dycheck, HyperNeRF)
% --> 주관적 화질 비교 그림 (줌인된 결과들로 3-4 scenes)
% Ablation studies
% Method +, -에 대한 table
% (a) 1stage: Implicit deformation only (Note: 1stage의 explicit only는 4DGS다)
% (b) G2L 2stage: 모두 Implicit
% (c) G2L 2stage: 모두 explicit (즉 4DGS를 2stage deformation 한거다)
% (d) G2L 2stage implicit + explicit
% (e) (e)에 learnable dynamics 적용
% (f) Temporal Adjustment까지 적용 (Ours)
% Analysis 1: 8개 segment로 나눈 구조에서, 각 Local CS가 모든 프레임으로 deform되도록. 자기가 전담하는 temporal interval을 가장 잘 표현하더라라는걸 graph로 보여줌
% Analysis 2: 8개 segment로 나눈 scene 중 하나에 대해서, optical flow magnitude와 TIA 결과가 얼마나 일치하는 지 도시
%------------------------------------ table start -----------------------------------%

\begin{table*}[!h]
\begin{center}
\setlength\tabcolsep{5pt} % default value: 6pt
\renewcommand{\arraystretch}{1.1}
\scalebox{0.8}{
\begin{tabular}{ c  |cc cc cc cc}
\bottomrule \hline
\noalign{\smallskip}
Method & \multicolumn{2}{c}{Apple}& \multicolumn{2}{c}{Block}& \multicolumn{2}{c}{Paper-windmill}& \multicolumn{2}{c}{Space-out}\\  
\hline\noalign{\smallskip}

SC-GS~\cite{Huang2024SCGS}    & 14.96 / 0.692 / 0.508 & 173.3& 13.98 / 0.548 / 0.483& 115.7& 14.87 / \best{0.221} / 0.432 & 446.3& \best{14.79} / 0.511 / \best{0.440} & 114.2\\

Deformable 3DGS ~\cite{Yang2024Deformable3DGS}       & \second{15.61} / \second{0.696} / \best{0.367}& 87.71 & \second{14.87} / \second{0.559} / \best{0.390}& 118.9& \second{14.89} / 0.213 / \best{0.341}& 160.2& 14.59 / 0.510 / \second{0.450}& \second{42.01}\\

4DGS~\cite{Wu20244DGS}   & 15.41 / 0.691 / 0.524& \second{61.52}& 13.89 / 0.550 / 0.539& \second{63.52}& 14.44 / 0.201 / 0.445& 123.9& 14.29 / \second{0.515} / 0.473& 52.02\\

\textbf{MoDec-GS (Ours)}& \best{16.48} / \best{0.699} / \second{0.402} & \best{23.78} 
& \best{15.57} / \best{0.590} / \second{0.478}& \best{13.65} 
& \best{14.92} / \second{0.220} / \second{0.377}& \best{17.08} 
& \second{14.65} / \best{0.522} / 0.467& \best{18.24}
\end{tabular}
}

\scalebox{0.8}{
\begin{tabular}{ c  |cc cc cc |cc}
\hline\noalign{\smallskip}
 & \multicolumn{2}{c}{Spin}& \multicolumn{2}{c}{Teddy}& \multicolumn{2}{c}{Wheel}& \multicolumn{2}{c}{\textbf{Average}}\\  
\hline\noalign{\smallskip}

SC-GS~\cite{Huang2024SCGS}    & 14.32 / 0.407 / 0.445& 219.1& \second{12.51} / \second{0.516} / \best{0.562}& 318.7& \second{11.90} / \second{0.354} / 0.484 & 239.2& \second{13.90} / \second{0.464} / 0.479& 232.4\\

Deformable 3DGS ~\cite{Yang2024Deformable3DGS}    & 13.10 / 0.392 / 0.490& 133.9& 11.20 / 0.508 / \second{0.573}& 117.1& 11.79 / 0.345 / \best{0.394}& 106.1& 13.72 / 0.461 / \best{0.430}& 109.4\\

4DGS~\cite{Wu20244DGS}  & \second{14.89} / \second{0.413} / \second{0.441} & \second{71.80} & 12.31 / 0.509 / 0.605& \second{80.44}& 10.83 / 0.339 / 0.538& \second{96.50}& 13.72 / 0.460 / 0.509& \second{78.54}\\

\textbf{MoDec-GS (Ours)}            & \best{15.53} / \best{0.433} / \best{0.366}& \best{26.84} & \best{12.56} / \best{0.521} / 0.598& \best{12.28} & \best{12.44} / \best{0.374} / \second{0.413}& \best{16.68} & \best{14.60} / \best{0.480} / \second{0.443}& \best{18.37}\\

\bottomrule \hline

\end{tabular}
}
\vspace{-2mm}
\caption{\textbf{Quantitative results comparison on the iPhone datasets \cite{Gao2022Dycheck}}. \textcolor{red}{\textbf{Red}} and \textcolor{blue}{\underline{blue}} denote the best and the second best performances, respectively. Each block element of 4-performance denotes (mPSNR(dB)$\uparrow$ / mSSIM$\uparrow$ / mLPIPS$\downarrow$ \, Storage(MB)$\downarrow$).} 

\label{table:iphone_comparison}
\end{center}
\vspace{-0.5cm}
\end{table*}

%------------------------------------ table end -----------------------------------%

\begin{figure*}[!h]
\centering
\includegraphics[scale=0.54]{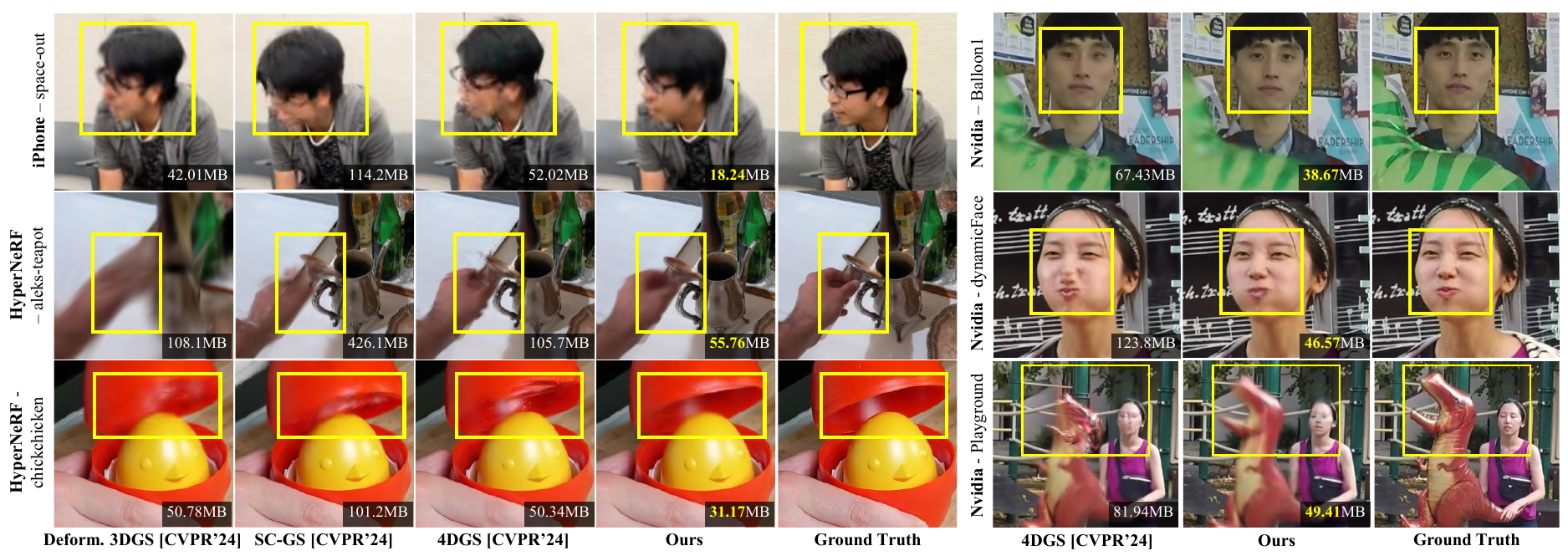}
\vspace{-2mm}
\caption{\textbf{Qualitative results comparison on three datasets \cite{Gao2022Dycheck, Park2021HyperNeRF, yoon2020novel}.} The yellow boxes highlight areas where the proposed method achieves notable visual quality improvements, and the storage for the corresponding sequence is displayed below each rendered patch.}
\label{fig:qualitative results}
\vspace{-0.5cm}
\end{figure*}

\vspace{-0.3cm}
\section{Experiments}
\subsection{Experimental Setup}
\label{subsec:experimental setup}

\noindent \textbf{Implementation Details.}\quad Our framework is built upon 4DGS \cite{Wu20244DGS} and Scaffold-GS \cite{lu2024scaffold}, retaining most hyperparameters. The parameters related to newly designed modules are empirically derived (see \textit{Suppl.} \ref{sub:Implementation details} for the details).

\noindent \textbf{Datasets and Metrics. }\quad 
There are various and well-validated multi-view videos datasets \cite{sabater2017dataset, li2022neural}, but achieving high rendering quality especially in \textit{monocular} reconstruction remains \textit{challenging}, which is the most accessible real-life application due to the prevalence of camera-equipped mobile devices. To focus on the complex motion present in such real-world videos, we evaluate our method on recent monocular video benchmark, Dycheck-iPhone \cite{Gao2022Dycheck}, which closely reflects the real-life video characteristics without teleporting. We also used HyperNeRF \cite{Park2021HyperNeRF} and Nvidia-monocular dataset \cite{yoon2020novel}, which are widely used for monocular evaluation, employed to assess the generalization performance of our method.
For all three datasets, initial point cloud data was manually generated by COLMAP \cite{schonberger2016structure} using the script provided in \cite{Wu20244DGS}. 
The image quality of our approach is evaluated using three metrics: Peak Signal-to-Noise Ratio (PSNR), Structural Similarity Index (SSIM), and Learned Perceptual Image Patch Similarity (LPIPS) \cite{zhang2018unreasonable}. Each metric is computed per frame and subsequently averaged across all test frames. Storage efficiency is measured in megabytes (MB) by summing the size of the trained model. For iPhone dataset, we used the masked metrics based on the official co-visible mask provided by \cite{Gao2022Dycheck}. 

\noindent \textbf{Comparison Methods.}\quad  We compare our approach with recent dynamic 3D Gaussian representation methods that can reconstruct dynamic scenes from a monocular video footage: Deformable-3DGS \cite{Yang2024Deformable3DGS}, SC-GS \cite{Huang2024SCGS}, and 4DGS \cite{Wu20244DGS}. The official codes are used for the methods, and we adjusted several parameters to achieve reasonable rendering quality, with details provided in the \textit{Suppl.} \ref{sub:Implementation details}.
% 아래 포함, 보다 detail한 내용은 supple로 이동

\subsection{Results}
\noindent \textbf{Quantitative Comparison.}\quad As detailed in Tab. \ref{table:iphone_comparison}, our method significantly reduces storage while achieving the best or second-best visual quality performance across almost all sequences of iPhone dataset \cite{Gao2022Dycheck}. On average, it maintains or even improves visual quality with only about $6\%$ of the storage compared to the second-best method in terms of quality, SC-GS \cite{Huang2024SCGS}. On HyperNeRF dataset \cite{Park2021HyperNeRF}, we present only the average performance shown as Tab. \ref{table:hypernerf_nvidia}-(a). For this dataset, our method attains the best performance in PSNR, SSIM and second-best in LPIPS, while having approximately $18\%$ of SC-GS's storage \cite{Huang2024SCGS} and around $57\%$ of 4DGS's \cite{Wu20244DGS}. For the Nvidia dataset \cite{yoon2020novel}, we compare our method only with the second-best method in terms of visual quality relative to storage, 4DGS \cite{Wu20244DGS}. Tab. \ref{table:hypernerf_nvidia}-(b) shows that our method reduces storage while simultaneously improving visual quality in all metrics. 
Supplementary materials includes all per-sequence results (Tab. \ref{table:sub_all_results}), additional results on synthetic (Tab. \ref{table:D-NeRF}) and real-world (Tab. \ref{table:panoptic_sports}) datasets, and comparison with NeRF-extension frameworks (Tab. \ref{table:Nerf_comp}). Please refer to it for further details.

\noindent \textbf{Qualitative Comparison.}\quad To evaluate the visual quality of the proposed method, we conducted qualitative assessments on three datasets \cite{Gao2022Dycheck, Park2021HyperNeRF, yoon2020novel} shown in Fig. \ref{fig:qualitative results}. We focused particularly on regions where dynamic objects are in motion. As introduced in Fig. \ref{fig:figure_page1}, while comparison methods struggle with handling complex motion, our method demonstrates better quality in regions with such motion, thanks to GLMD. Furthermore, as shown in the NVIDIA dataset results, our method maintains fine visual quality even when static objects exhibit only local changes (e.g., facial expressions). This is due to TIA effectively localizing the coverage of Local CS. Not only does our method achieve these quality improvements, but it also maintains storage requirements at about half the average size of compared methods. 

%To validate the effectiveness of the proposed technique through qualitative evaluation, we observed the following two cases: where there is a global movement at the object level, and where there are local changes within a stationary object. Fig. \ref{fig:qualitative results} illustrates the former case. In HyperNeRF's \textit{cut-lemon} dataset\cite{Park2021HyperNeRF}, there is temporal segment where the knife object and the resulting cut lemon piece move globally. Focusing on this segment and comparing with other methods, our method shows fewer residual 3DGs between the moving spaces of the object, resulting in clearer visual quality without artifacts. Meanwhile, \skwak{Fig. \ref{fig:Local_movement_nvidia}} presents a qualitative evaluation of the case where an object locally varies without global movement. 

%------------------------------------ table 2 -----------------------------------%
\begin{table}[t]
\small
\begin{center}
\setlength\tabcolsep{5pt} % default value: 6pt
\renewcommand{\arraystretch}{1.3}
\scalebox{0.75}{
\begin{tabular}{ c | c c c c }
\bottomrule
\hline\noalign{\smallskip}
 & \multicolumn{4}{c}{(a) HyperNeRF}\\
Methods & PSNR$\uparrow$ & SSIM$\uparrow$ & LPIPS$\downarrow$ & Storage$\downarrow$ \\  
\hline\noalign{\smallskip}
SC-GS [CVPR'24] ~\cite{Huang2024SCGS}  & 26.95 & \second{0.815} & \best{0.213} & 226.0 \\
Deformable 3DGS [CVPR'24]  ~\cite{Yang2024Deformable3DGS}  & 25.96  & 0.766 & 0.294 & 87.13 \\
4DGS [CVPR'24] ~\cite{Wu20244DGS} & \second{27.44} & 0.797 & 0.302 & \second{72.65}\\
Ours & \best{27.78} & \best{0.827} & \second{0.219} & \best{40.82} \\
\bottomrule
\hline\noalign{\smallskip}
\end{tabular}}
\scalebox{0.75}{
\begin{tabular}{ c | c c c c }
 & \multicolumn{4}{c}{(b) Nvidia}\\
Methods & PSNR$\uparrow$ & SSIM$\uparrow$ & LPIPS$\downarrow$ & Storage$\downarrow$ \\  
\hline\noalign{\smallskip}
\;\;\qquad 4DGS [CVPR'24] ~\cite{Wu20244DGS} \qquad\;& 25.82 & 0.844 & 0.219 & 67.44 \\
Ours & \best{26.65} & \best{0.876} & \best{0.171} & \best{39.64} \\
\bottomrule
\hline\noalign{\smallskip}
\end{tabular}}

\end{center}
\vspace{-0.63cm}
\caption{\textbf{Quantitative results comparison on (a) HyperNeRF \cite{Park2021HyperNeRF} and (b) Nvidia monocular \cite{yoon2020novel} dataset}. }
\label{table:hypernerf_nvidia}
\vspace{-0.3cm}
\end{table}

%------------------------------------ table 4 -----------------------------------%
\begin{table}[t]
\small
\begin{center}
\setlength\tabcolsep{4pt} % default value: 6pt
\renewcommand{\arraystretch}{1.3}
\scalebox{0.7}{
\begin{tabular}{ l | c c c c }
\bottomrule
\hline\noalign{\smallskip}
Variant & mPSNR$\uparrow$ & mSSIM$\uparrow$ & mLPIPS$\downarrow$ & Storage$\downarrow$ \\  
\bottomrule
\hline\noalign{\smallskip}
(a) 1stage, Gaussian deform (\cite{Wu20244DGS}) & 13.73 & 0.460 & 0.509 & 78.54 \\
(b) 1stage, anchor deform & 13.56  & 0.449 & 0.510 & \colorbox{GreenYellow}{36.92} \\
(c) 2stage, all anchor deform & 13.93 & 0.453 & 0.492 & 55.29\\
(d) 2stage, GAD + LGD (\textbf{GLMD}) & 14.48 & 0.475 & 0.455 & 49.70\\
(e) (d) with smaller hexplane & 14.46 & 0.475 & 0.451 & \colorbox{GreenYellow}{\second{22.67}}\\
(f) (e) with $d_G$ and $d_L$ (anchor dynamics) & \second{14.51} & \second{0.478} & \second{0.447} & 22.72 \\
(g) (f) with \textbf{TIA}  (our final MoDec-GS)& \best{14.60} & \best{0.480} & \best{0.443} & \colorbox{GreenYellow}{\best{18.37}} \\
\bottomrule
\hline\noalign{\smallskip}
\end{tabular}}
\end{center}
\vspace{-0.63cm}
\caption{\textbf{Ablation studies on MoDec-GS components}. Each row evaluates the impact of a specific design choice. Yellow-green cells highlight configurations with substantial storage reduction.}
%Yellow-green backgrounds highlight cases where the applying of the proposed method resulted in a noticeable reduction in storage.}
\label{table:ablation_study}
\vspace{-0.5cm}
\end{table}

\subsection{Analysis}

\begin{figure}[!t]
\centering
\includegraphics[scale=0.32]{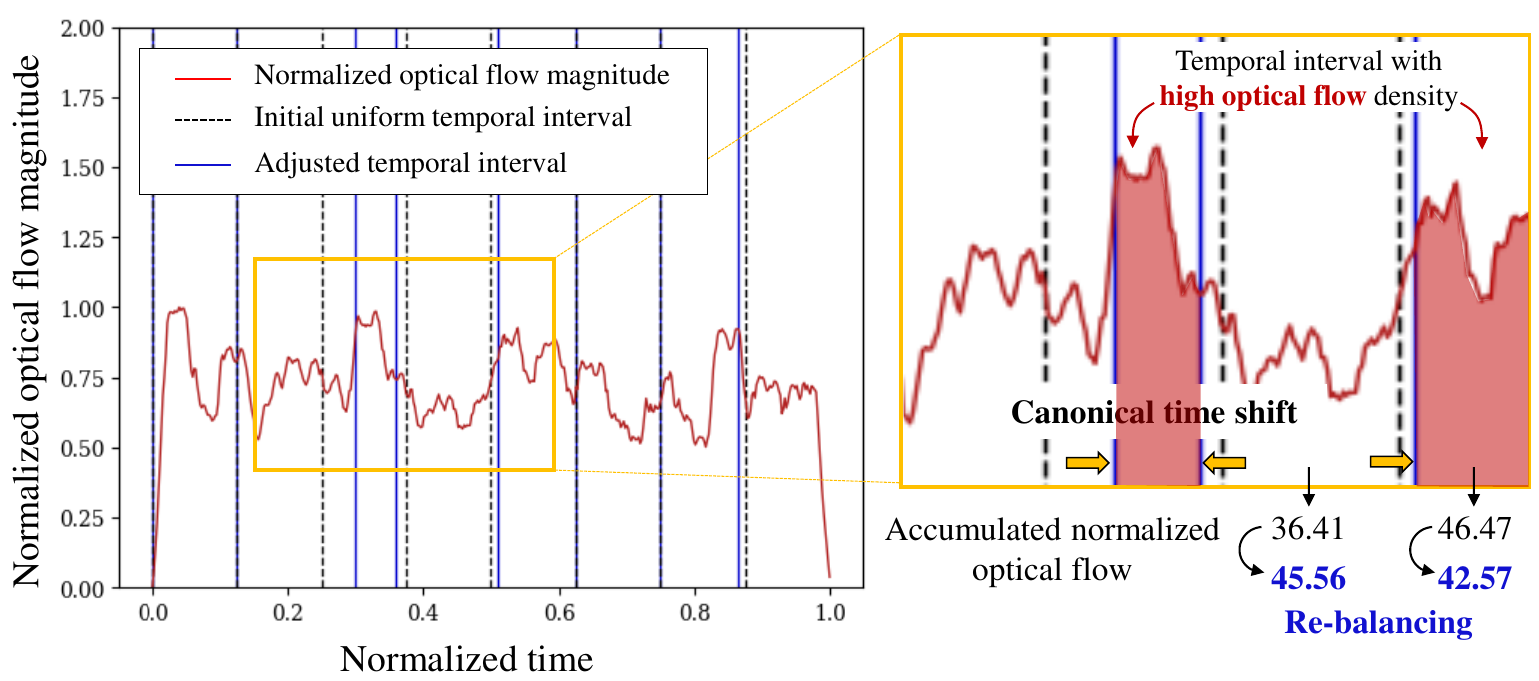}
\vspace{-0.5cm}
\caption{\textbf{Effectiveness of TIA.} Initially uniform intervals (black dotted lines) are adaptively reallocated based on motion complexity (blue lines), as indicated by normalized optical flow magnitude.}
\label{fig:TIA validation}
\vspace{-0.5cm}
\end{figure}

\noindent \textbf{Ablation studies.} \quad We analyze the effectiveness of the components in our MoDec-GS through comprehensive ablation studies as shown in Tab \ref{table:ablation_study}. All results are averaged over all the iPhone sequences \cite{Gao2022Dycheck}. Note that our baseline is a single-stage deformation method that explicitly deforms Gaussians, as in \cite{Wu20244DGS}. We first examine the effectiveness of our anchor deformation - (b). 
Leveraging anchor-based representation \cite{lu2024scaffold} slightly reduces performance but significantly cuts storage (52$\%$ reduction). 
Configuring the Local CS by adding a global hexplane - (c), we observe that the performance improvement outweighs the increase in storage due to the additional grid, compared to (a). We then show the effectiveness of LGD - (d). 
Instead of applying consistent anchor deformation for both global and local deformation, performing LGD after reconstructing the neural Gaussian noticeably improves performance. 
%Additionally, it allows for a slight reduction in storage, because the size of Global CS can be decreased through anchor adjustment, as the regions represented by adjacent Gaussians no longer require anchors.
It also reduces storage by allowing smaller Global CS via anchor adjustment, as the regions represented by adjacent Gaussians no longer require anchors. 
We further analyzed GLMD's handling of complex motion via optical flow in \textit{Suppl}. \ref{sup:visualization of GLMD} and \ref{Analysis of Complex Motion through GLMD}. 
The efficient design of GLMD allows for a reduction in the size of the global and local hexplanes with minimal impact on visual quality - (e) (55$\%$ reduction). 
By adding the proposed learnable parameters that control anchor's motion dynamics, the increase is almost negligible while improving visual quality to some extent. Finally, applying TIA to this variant - our final MoDec-GS - enables both quality improvement and storage reduction. When temporal intervals are appropriately adapted to the degree of motion in the scene through TIA, the representational capacity of the limited-size hexplane can be utilized more efficiently, and also the Gaussian movements represented by each LGD become simpler (See Fig. \ref{fig:2stage_deformation}), which further enhances the efficiency of anchor deformation. 
Additional experiments verified TIA's effects. We precomputed optical flow  \cite{shi2023flowformer++} to measure the degree of motion in the scene, and evaluated how the TIA responds to it. As shown in Fig. \ref{fig:TIA validation}, initially uniform temporal intervals (black dotted line) are adjusted into non-uniform intervals (blue solid line) during the training process, shrinking and shifting toward regions with relatively higher normalized optical flow magnitude. Examining accumulated flow confirms TIA effectively \textit{rebalances} the degree of motion across intervals.
Overall, each additional component in MoDec-GS contributes to either improved visual fidelity or enhanced memory efficiency, with our final configuration achieving the best balance.

%% file: sec/6_conclusion.tex
\vspace{-2mm}
\section{Conclusion}
\label{sect:conclusion}
\vspace{-2mm}
We propose MoDec-GS, a novel compact framework for high-quality dynamic 3D Gaussian splatting, addressing storage demands and complex motion challenges in dynamic scene reconstruction. By utilizing Global-to-Local Motion Decomposition (GLMD), which incorporates Global Anchor Deformation (GAD) for global motion and Local Gaussian Deformation (LGD) for fine-grained local adjustments, MoDec-GS effectively captures complex motions while minimizing storage use. Additionally, our Temporal Interval Adjustment (TIA) allows adaptive temporal segmentation, across dynamic intervals without requiring external motion data. Extensive evaluations confirm that MoDec-GS significantly reduces model size—up to average 70$\%$—while either preserving or enhancing rendering quality across challenging datasets, offering a compact yet powerful solution for real-world dynamic 3D reconstruction.

%% file: sec/X_suppl.tex
\clearpage
\setcounter{page}{1}
\maketitlesupplementary
\appendix

% main paper와 supple의 renference 다를 수 있다. 여기 달린거로 봐라. plase note that.. 

%Please note that the reference numbers in the main paper and supplementary material are not aligned. Please see the reference numbers in the supplementary material. 

\section{Project Page and Demo Video}
\label{sub:demo videos}

Please refer to our project page: \url{https://kaist-viclab.github.io/MoDecGS-site/}. The project page provides a summarized description of our method, an interactive visual comparison demo, and demo videos. In the demo video (\url{https://youtu.be/5L6gzc5-cw8}), we demonstrated subjective quality comparisons for HyperNeRF's interp-cut-lemon, interp-torchocolate, misc-espresso, misc-tampling, vrig-peel-banana. We compared four framworks which are SC-GS \cite{Huang2024SCGS}, Deformable 3DGS \cite{Yang2024Deformable3DGS}, 4DGS \cite{Wu20244DGS}, and MoDec-GS (Ours). The videos are concatenated in a 2$\times$2 or 4$\times$1 format depending on the shape of the video. Additionally, the code will be released through a GitHub repository: \url{https://github.com/skwak-kaist/MoDec-GS}.

\section{Implementation Details}

\label{sub:Implementation details}
MoDec-GS is implemented using PyTorch and built upon 4DGS \cite{Wu20244DGS} and Scaffold-GS \cite{lu2024scaffold}
codebases. Similar to 4DGS \cite{Wu20244DGS} we adopt a hexplane-based deformation method to represent video content, while using an anchor-based representation \cite{lu2024scaffold} for the canonical 3D Gaussians. The key hyperparameters for the anchor repesentation include n\_offset=20, voxel\_size=0.01, feat\_dim=32, and appearance\_dim=16, with no feature bank utilized. Iterations are set as follows: 3,000 for the Global stage, and between 20,000 and 60,000 for the Local stage depending on the sequence length. The global and local hexplanes are set to $[32, 32, 32, 10]$ and $[64, 64, 64, 100]$ with a two-level multi-resolution, respectively, for all test cases. The parameters for TIA are as follows: $T^{\text{TIA}}_{\text{from}}$ is set to 500, $T^{\text{TIA}}_{\text{until}}$ is set to 10,000 or 20,000 depending on the total number of iterations, and $T^{\text{TIA}}_{\text{period}}$ is set to 1,000, also depending on the iterations. $\tau_{\text{TIA}}$ is set to either 1.0 or 1.5, and the $s_{\text{TIA}}$ is chosen within the range of 0.01 to 0.1. 
For the comparison methods, Deformable-3DGS and SC-GS were compared in the same local experimental environment and retrained on all datasets. For the Deformable-3DGS, the option for 6-degrees of freedom deformation is turned on for better rendering quality. We set the number of node and the dimension of hyper coordinates for the SC-GS at 2048 and 8, respectively. No mask images for background separation were used. 

\section{Preliminaries}
\subsection{Anchor-based representation}
\label{sub:anchor-based representation}

\begin{figure*}[t!]
\centering
\includegraphics[scale=0.6]{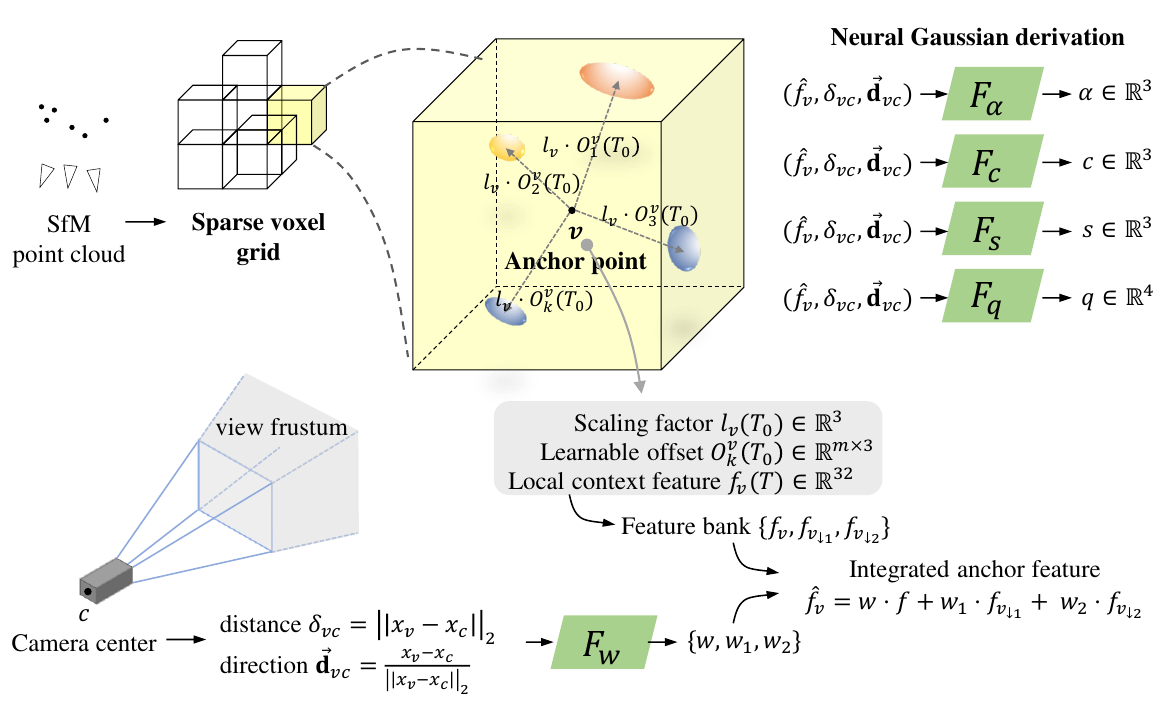}
\vspace{-0.2cm}
\caption{\textbf{Overview of the anchor-based 3DGS representation process} \cite{lu2024scaffold}.}
\vspace{-0.2cm}
\label{fig:sub_fig4}
\end{figure*}

 Lu \textit{et al}. \cite{lu2024scaffold} proposed Scaffold-GS, a structured anchor-based 3DGS representation approach, designed to improve efficiency, robustness and scalability in novel view synthesis. Unlike traditional 3DGS, which often results in redundant Gaussians due to excessive fitting to training views, this approach introduces a hierarchical and structured representation. The method begins by initializing a sparse set of anchor points from Structure-from-Motion (SfM) points and distributing neural Gaussians around these anchors. As shown in Fig. \ref{fig:sub_fig4}, each anchor point is associate with learnable offsets and a scaling factor, allowing local Gaussians to be dynamically placed and adapted to varying viewpoints. Therefore, instead of allowing Gaussians to drift freely like in 3DGS \cite{kerbl20233dgs}, Scaffold-GS constrains their placement using anchor points. Given an anchor at position $x_v$, the position of $k$ derived Gaussians are computed as: 
 \begin{equation} \label{eq:positional_offset_scaffold_gs}
  \{\mu_0,\cdots,\mu_{k-1}\}=x_v + \{\textit{O}_0,\cdots,\textit{O}_{k-1}\}\cdot l_v,
\end{equation}
 where $\textit{O}_i$ are learnable offsets and $l_v$ is a scaling factor. Their opacity values $\alpha$, colors $c$, and other attributes are decoded through MLPs based on the local context feature and view-dependent information: 
 \begin{equation} \label{eq:alpha_decoding_scaffold_gs}
  \{\alpha_0,\cdots,\alpha_{k-1}\} = F_{\alpha}(f_v, \delta_{vc}, \textbf{d}_{vc}),
\end{equation}
 where $\delta_vc$ and $\textbf{d}_{vc}$ represent the distance and direction from the camera to the anchor. The attributes of these Gaussians - position, opaicty, color and scale - are predicted on-the-fly based on local context feature assigned on the anchor and the viewing direction. This view-adaptive mechanism prevents excessive redundancy and enhances robustness to complex scene structures. 
  To further refine the representation, Scaffold-GS employs a growing and pruning strategy. Growing introduces new anchors in underrepresented areas where the gradient magnitude of neural Gaussians exceeds a predefined threshold. Pruning removes anchors that consistently produce low-opacity Gaussians, ensuring an efficient representation. At inference time, only Gaussians within the view frustum and with significant opacity contribute to rendering, maintaining real-time performance.

\subsection{HexPlane-based deformation encoder}
In this work, we employ the HexPlane which is widely adopted for the deformation fields in dynamic scene representations \cite{Wu20244DGS, keil2023Kplanes}. The spatio-temporal flow can be predicted by inputting the feature vector that corresponds to each spatio-temporal coordinate $(x, y, z, t)$, the resulting hexplane feature vector $H_{L}$ can be expressed as follows:
\vspace{-1mm}
\begin{equation} \label{eq:hexplane_feature_hadamard_multiscale}
    H_{L}(x,y,z,t) = \bigcup_{s \in S} \prod_{p \in P} f(x,y,z,t)^{s}_{p},
\end{equation}
\noindent where $f(x,y,z,t)^{s}_{c}$ denote the interpolated feature corresponding to the queried four-dimensional coordinate, $p$ and $s$ are the indices for the two-dimensional planes and the grid scales, respectively.

\section{Concurrent Works}
\label{sup: concurrent}
%Very recently, a concurrent work \cite{cho20244d} proposes a method to extend 3D scaffolds \cite{lu2024scaffold} to 4D space for compactly representing dynamic scenes. However, this approach focuses solely on multi-view scenes and is not capable to handle casual monocular data, which is a more common real-world setting. Additionally, it assumes object motion to be piecewise linear within uniformly divided time intervals, lacking a solution for handling complex motions, which generally contains non-linear motions. %In contrast, our MoDec-GS can accommodate both multi-view and monocular content, providing an effective solution for dynamic scenes with complex motion where global and local motions are combined over non-uniform time intervals. 
%Another concurrent work, Relay-GS, has been proposed to effectively represent large-scale complex motion through temporal segment partitioning. This study first learns the fundamental scene structure from all frames and then decouples the highly dynamic foreground using a learnable mask. Subsequently, it blends the start, middle, and end points of each segment to generate pseudo-views, where moving semi-transparent foreground Gaussians, referred to as Relay Gaussians, are positioned. However, this study is also not applicable to monocular views, as it cannot determine the camera paramters for pseudo-views. 
Very recently, some concurrent works \cite{cho20244d, gao2024relaygs}, have been proposed. Cho \textit{et al.} \cite{cho20244d} proposed a framework for extending 3D scaffolds into 4D space, aiming to efficiently represent 4D Gaussians through the introduction of neural velocity-based time-variant Gaussians and temporal opacity. Another work, Relay-GS \cite{gao2024relaygs}, proposed a framework for effectively handling large-scale complex motion by modeling motion within temporal segments. It utilizes a learnable mask to separate the dynamic foreground and employs pre-generated pseudo-views, where semi-transparent Gaussians—\textit{Relay Gaussians}—are placed along the trajectory. Both studies propose compact and efficient dynamic Gaussian representations for real-world scenes. However, they share a common limitation: neither can handle \textit{casual monocular data}, which is closer to real-world settings.

\section{Comparison to NeRF-extension Methods}
\label{sub:NeRF-extension}

\begin{figure*}[!h]
\centering
\includegraphics[scale=0.75]{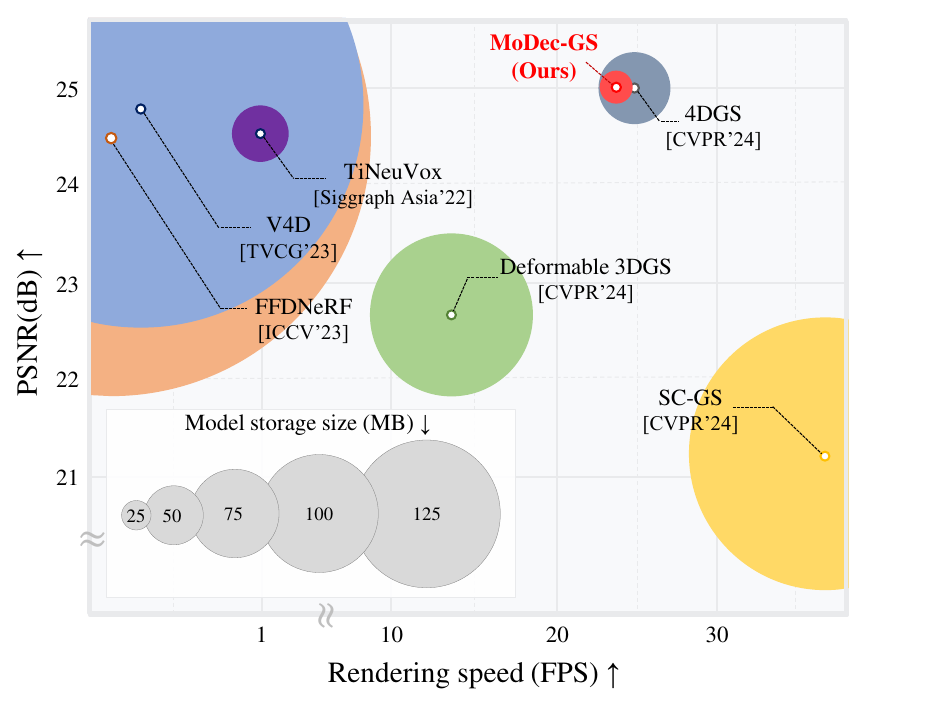}
\vspace{-0.2cm}
\caption{\textbf{Performance comparison visualization graph.} The $x$-axis represents rendering speed (FPS)$\uparrow$, and the $y$-axis indicates PSNR$\uparrow$. Each framework is depicted as a bubble, with the size of the bubble representing the model storage size (MB)$\downarrow$.}
\label{fig:sub_fig3}
\end{figure*}

Recent trends in NVS are driven by 3DGS \cite{kerbl20233dgs} and its extensions \cite{liu2024compgs, sun20243dgstream, Wu20244DGS, Huang2024SCGS, Lee2024Compact3DGS, Yang2024Deformable3DGS}. However, NeRF-based methods that utilize differentiable volume rendering are still being actively researched and demonstrate strong performance in terms of visual quality \cite{Muller2022InstantNGP, keil2023Kplanes, liu2020neural, kratimenos2025dynmf, liu2023robust}. Although our primary target application focuses on being \textit{compact} without losing \textit{real-time rendering} capabilities, we also provide a comparison with NeRF-based approaches for reference information using results taken from \cite{Wu20244DGS}. Tab. \ref{table:Nerf_comp} presents the comparison results. In the table, the numbers for \cite{park2021nerfies, Park2021HyperNeRF, fang2022fast, kerbl20233dgs, guo2023forward} are sourced from \cite{Wu20244DGS} and those of the other are generated in our local environment. We confirmed that the performances of 4DGS reported in \cite{Wu20244DGS} is nearly reproduced in our side, and note that there are differences of GPU environment (\cite{Wu20244DGS}: RTX 3090, Ours: RTX A6000 - due to the time limitation, the speed comparison measured on the same machine has not prepared, but it is generally known that RTX A6000 is slower than RTX 3090. We plan to fairly measure the training time and rendering speed on the same GPU in the future). Through the comparison, we confirmed that our method achieved the lowest storage requirement at only 52$\%$ of the second-best \cite{fang2022fast}, while maintaining the highest visual quality scores and high-speed rendering performance exceeding 20 fps. Additionally, to visualize the comparison results with other frameworks, we present a performance comparison graph, as shown in Fig. \ref{fig:sub_fig3} where the $x$-axis represents rendering speed (FPS), the $y$-axis denotes PSNR, and the bubble size (MB) indicates the model's storage size. 
Our method achieves an exceptionally small storage size while maintaining the highest level of visual quality performance.

%------------------------------------ Speed -----------------------------------%
\begin{table*}[!h]
\small
\begin{center}
\setlength\tabcolsep{4pt} % default value: 6pt
\renewcommand{\arraystretch}{1.3}
\scalebox{0.85}{
\begin{tabular}{ l | c c | c | c c }
\bottomrule
\hline\noalign{\smallskip}
Methods & PSNR(dB)$\uparrow$ & MS-SSIM$\uparrow$ & Training times$\downarrow$ & Run times(FPS)$\uparrow$ & Storage(MB)$\downarrow$ \\  
\bottomrule
\hline\noalign{\smallskip}
Nerfies \cite{park2021nerfies} & 22.2 & 0.803 & $\sim$ hours & $<$ 1 & - \\
HyperNeRF \cite{Park2021HyperNeRF} & 22.4 & 0.814 & 32 hours & $<$ 1 & - \\
TiNeuVox-B \cite{fang2022fast}& 24.3 & 0.836 & \best{30 mins} & 1 & \second{48} \\
FFDNeRF \cite{guo2023forward}& 24.2 & \best{0.842} & - & 0.05 & 440 \\
V4D \cite{gan2023v4d} & 24.8 & 0.832 & 5.5 hours & 0.29 & 377 \\
\hline\noalign{\smallskip}
3DGS \cite{kerbl20233dgs}& 19.7 & 0.680 & \second{40 mins} & \best{55} & 52 \\
4DGS \cite{Wu20244DGS} & \second{25.0} & \second{0.838} & 1.2 hour & \second{24.9} & 61 \\
\textbf{MoDec-GS (Ours)} & \best{25.0} & 0.836 & 1.2 hour & 23.8 & \best{28} \\

\bottomrule
\hline\noalign{\smallskip}
\end{tabular}}
\end{center}
\vspace{-4mm}
\caption{\textbf{Performance comparison with a NeRF-extension framework, including training and rendering speed.} Averaged over 536$\times$960 HyperNeRF's vrig datasets \cite{Park2021HyperNeRF}. The performance numbers of \cite{park2021nerfies, Park2021HyperNeRF, fang2022fast, kerbl20233dgs, guo2023forward} are sourced from \cite{Wu20244DGS}. The training times and run times reported in \cite{Wu20244DGS} were measured on an NVIDIA RTX 3090 GPU, while our framework was tested on an RTX A6000 GPU. Please note that the A6000 GPU has approximately 20 $\%$ lower memory bandwidth compared to that of the RTX 3090.} 
\label{table:Nerf_comp}
\end{table*}

\section{Detailed Experimental Results}

\subsection{Datasets and metrics}
In this paper, the following three datasets are used: iPhone \cite{Gao2022Dycheck}, HyperNeRF \cite{Park2021HyperNeRF}, and Nvidia \cite{yoon2020novel}. All datasets were downloaded from their official repositories, and the COLMAP \cite{schonberger2016structure} data for iPhone and HyperNeRF datasets were directly generated using the script provided in \cite{Wu20244DGS} by ourselves. Note that the script is designed to subsample frames to ensure that the number of frames does not exceed 200 when obtaining the initial point cloud. For COLMAP inputs, we used the 2$\times$ version of the sequences for each dataset. Specifically, 360$\times$480 for the iPhone dataset and 536{$\times$}960 for the HyperNeRF dataset. For the Nvidia dataset, multi-view video frames were sampled sequentially at one frame per timestamp, resulting in a total of 192 \textit{monocular} frames. To define the test frames, every 8th frame was excluded from the training views. This is one of the settings provided in \cite{Wu20244DGS}, resulting in a total of 168 training views and 24 test views meaning the temporal interpolation, which is more challenging setting. Through validation on this setting of NVIDIA dataset, which features long-range time duration and high resolution (around FHD to 2K), we aimed to effectively verify the storage reduction capabilities of our model. 
Regarding the metrics, PSNR, SSIM \cite{wang2004image}, and LPIPS \cite{zhang2018unreasonable} metrics are calculated using the functions in \cite{Wu20244DGS}, while masked metrics for the iPhone dataset are obtained by the functions and covisible masks provided by DyCheck \cite{Gao2022Dycheck}. For tOF \cite{chu2020learning}, module form TecoGAN \cite{chu2018temporally} is utilized. 
For model storage, it is calculated as the sum of the sizes of a single global CS ply, two deformation fields, per-attribute MLPs, and the canonical time list. Note that $H_L$ is shared across temporal intervals, meaning that only a single pair of $H_G$ and $H_L$ exists.

\label{sub:Detailed_exp_results}
\subsection{Detailed results}
The full quantitative results on three datasets are presented in Tab \ref{table:sub_all_results}. Our method achieves the best or second-best visual quality performance in almost all sequences while using significantly less storage. Regarding the average performance of HyperNeRF, not only in the \textit{interp} results which are reported in the main paper, but also in \textit{misc} and \textit{vrig}, our method shows the highest PSNR/tOF and second-best SSIM performance, while using about 40$\%$ less storage compared to the second-best model from a storage perspective \cite{Wu20244DGS}.

\label{sub:dataset_generalization}
\subsection{Generalization to additional datasets}
We further evaluated our method's robustness using the D-NeRF \cite{pumarola2021d} and PanopticSports \cite{joo2015panoptic} datasets, each representing synthetic and real-world complex motion characteristics, respectively. For D-NeRF, we referenced the results form Compact Dynamic 3DGS (C. D. 3DGS) \cite{katsumata2024compact}. For PanopticSports, the results are adopted from TC-3DGS \cite{javed2024temporally}. As confirmed by the experimental results, our method demonstrate considerble rendering quality while maintaining low storage requirements, in both synthetic scenes and real-world complex motions.

%------------------------------------ D-NeRF -----------------------------------%
\begin{table}[h]
\small
\begin{center}
\setlength\tabcolsep{4pt} % default value: 6pt
\renewcommand{\arraystretch}{1.3}
\scalebox{0.8}{
\begin{tabular}{ l | c c c | c }
\bottomrule
\hline\noalign{\smallskip}
Methods & PSNR(dB)$\uparrow$ & MS-SSIM$\uparrow$ & LPIPS$\downarrow$ & Storage(MB)$\downarrow$ \\  
\bottomrule
\hline\noalign{\smallskip}
TiNeuVox-S \cite{fang2022fast} & 30.75 & 0.96 & 0.07 & \best{8} \\ 
TiNeuVox-B \cite{fang2022fast} & 32.67 & 0.97 & \second{0.04} & \second{48} \\ 
V4D \cite{gan2023v4d}       & \best{33.72} & \second{0.98} &\best{0.02} & 1200 \\
C.D.3DGS \cite {katsumata2024compact} & 32.19 & 0.97 & \second{0.04} & 159 \\
\hline\noalign{\smallskip}
\textbf{MoDec-GS (Ours)} & \second{33.25} & \best{0.99} & \best{0.02} & \best{8} \\

\bottomrule
\hline\noalign{\smallskip}
\end{tabular}}
\end{center}
%\vspace{-5mm}
\vspace{-3mm}
\caption{\textbf{Performance comparison on D-NeRF dataset.} The results were averaged over all sequences in the dataset, and the values for the comparison method were taken from \cite{katsumata2024compact}. }
\label{table:D-NeRF}
\end{table}

%------------------------------------ Panoptic Sports -----------------------------------%
\begin{table}[h]
\small
\begin{center}
\setlength\tabcolsep{4pt} % default value: 6pt
\renewcommand{\arraystretch}{1.3}
\scalebox{0.8}{
\begin{tabular}{ l | c c c | c }
\bottomrule
\hline\noalign{\smallskip}
Methods & PSNR(dB)$\uparrow$ & SSIM$\uparrow$ & LPIPS$\downarrow$ & Storage(MB)$\downarrow$ \\  
\bottomrule
\hline\noalign{\smallskip}
Dynamic 3DGS \cite{luiten2024dynamic} & \best{28.70} & \second{0.91} & 0.17 & 2008 \\
STG \cite{li2024spacetime} & 20.45 & 0.79 & \best{0.10} & \best{19} \\
4DGS \cite{Wu20244DGS} & 27.22 & \second{0.91} & \best{0.10} & 63 \\
TC-3DGS \cite{javed2024temporally} & 27.81 & 0.89 & 0.20 & 49 \\
\hline\noalign{\smallskip}
\textbf{MoDec-GS (Ours)} & \second{27.96} & \best{0.95} & \second{0.13} & \second{34} \\
\bottomrule
\hline\noalign{\smallskip}
\end{tabular}}
\end{center}
%\vspace{-5mm}
\vspace{-3mm}
\caption{\textbf{Performance comparison on PanopticSports dataset.} Results for the comparison method were sourced from \cite{javed2024temporally}.}
\label{table:panoptic_sports}
\end{table}

\section{Ablation Studies}
\label{sub:ablations studies}

\subsection{Rendering Overhead by 2-stage Deformation}
\label{sup:Computational Overhead by 2-stage Deformation}
As shown in Tab. \ref{table:Nerf_comp}, our method experiences only a marginal drop in FPS compared to 4DGS \cite{Wu20244DGS}, while maintaining real-time rendering capability. To further clarify the computational overhead introduced by the 2-stage deformation, we compared the rendering speed when using only a 1-stage deformation in our method. This corresponds to (b) in the ablation studies of Tab. \ref{table:ablation_study}. As in Tab. \ref{table:Nerf_comp}, the rendering speed comparison was conducted on the HyperNeRF's vrig dataset, and the results are presented in Tab. \ref{tab:rendering speed_stages}.

%------------------------------------ rendering speed -----------------------------------%
\begin{table}[!h]
\centering
\scalebox{0.8}{
\begin{tabular}{ c | c }
\bottomrule
\hline\noalign{\smallskip}
Method & Rendering speed (FPS) \\
\hline\noalign{\smallskip}
Ours (1-stage) & 24.7 \\
Ours (2-stage) & 23.8 \\
\bottomrule
\hline\noalign{\smallskip}
\end{tabular}}
\caption{\textbf{Rendering speed comparison} between 1-stage and 2-stage deformation of our method. }
\label{tab:rendering speed_stages}
\end{table}

\subsection{Hyperparameter studies}
We conducted an ablation study to assess the robustness of our framework and analyze the impact of hyperparameter variations. To align with the results in Tab. \ref{table:ablation_study}, we performed experiments by varying several key parameters on the iPhone \cite{Gao2022Dycheck} dataset. We conducted variation experiments on the local hexplane $H_L$ size, voxel size $\epsilon$, and the number of Gaussians per grid cell $N_{\text{offset}}$ as shown in Tab. \ref{tab:hyper-parameters ablation}. The default settings are shown in the middle column of the table. We observed that a trade-off between quality and storage depending on the  HexPlane/voxel grid resolution and the number of Gaussians per grid cell $N_{\text{offset}}$. The current setting provides a well-balanced compromise between these factors.

%------------------------------------ hyper parameter -----------------------------------%
\begin{table*}[!h]
    \centering
    \scalebox{0.8}{
    \begin{tabular}{c | ccc|c| ccc|c| ccc|c}
         \bottomrule
        \hline\noalign{\smallskip}
         Params &  \multicolumn{4}{c|}{Variation A}&  \multicolumn{4}{c|}{Default}&  \multicolumn{4}{c}{Variation B}\\
         & PSNR$\uparrow$ & SSIM$\uparrow$ & LPIPS$\downarrow$ & Storage$\downarrow$ 
         & PSNR$\uparrow$ & SSIM$\uparrow$ & LPIPS$\downarrow$ & Storage$\downarrow$
         & PSNR$\uparrow$ & SSIM$\uparrow$ & LPIPS$\downarrow$ & Storage$\downarrow$ \\
         \bottomrule
        \hline\noalign{\smallskip}
         $H_L$ size&  \multicolumn{4}{c|}{$[32, 32, 32, 50]$}&  \multicolumn{4}{c
         |}{$[64, 64, 64, 100]$}&  \multicolumn{4}{c}{$[128, 128, 128, 150]$}\\
         & 14.28 & 0.330 & 0.476 & 16.23 & 14.60 & 0.480 & 0.443 & 18.37 & 14.61 & 0.489 & 0.416 & 29.65 \\
         \hline\noalign{\smallskip}
         Voxel size &  \multicolumn{4}{c|}{$0.1$}&  \multicolumn{4}{c
         |}{$0.01$} &  \multicolumn{4}{c}{$0.001$}\\
         & 13.93 & 0.332 & 0.528 & 17.46 & 14.60 & 0.480 & 0.443 & 18.37 & 14.45 & 0.475 & 0.429 & 23.12 \\
         \hline\noalign{\smallskip}
         $N_{\text{offset}}$ & \multicolumn{4}{c|}{$5$}&  \multicolumn{4}{c
         |}{$10$} & \multicolumn{4}{c}{$20$}\\
         & 13.80 & 0.322 & 0.513 & 15.15 & 14.60 & 0.480 & 0.443 & 18.37 & 14.54 & 0.486 & 0.422 & 23.00 \\
          \bottomrule
        \hline\noalign{\smallskip}
    \end{tabular}
    }
    \caption{\textbf{Hyper-parameter variation experiments} on the local hexplane size, voxel size, and the number of Gaussians per grid cell. The default settings used in the main paper's experiments are shown in the middle column. }
    \label{tab:hyper-parameters ablation}
\end{table*}

\begin{figure*}[h]
\centering
\includegraphics[scale=0.5]{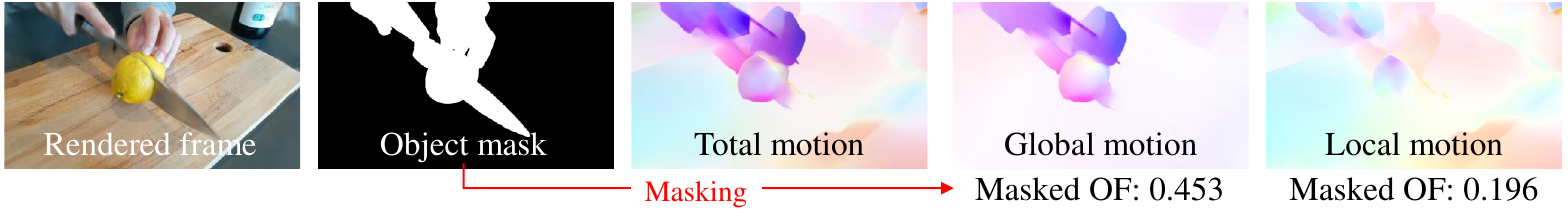}
\caption{\textbf{Masked optical flow analysis for GLMD.}}
\label{fig:Optical flow analysis for GLMD}
\vspace{-0.5cm}
\end{figure*}

\subsection{Visualization of GLMD}
\label{sup:visualization of GLMD}
\begin{figure*}[h!]
\centering
\includegraphics[scale=1.0]{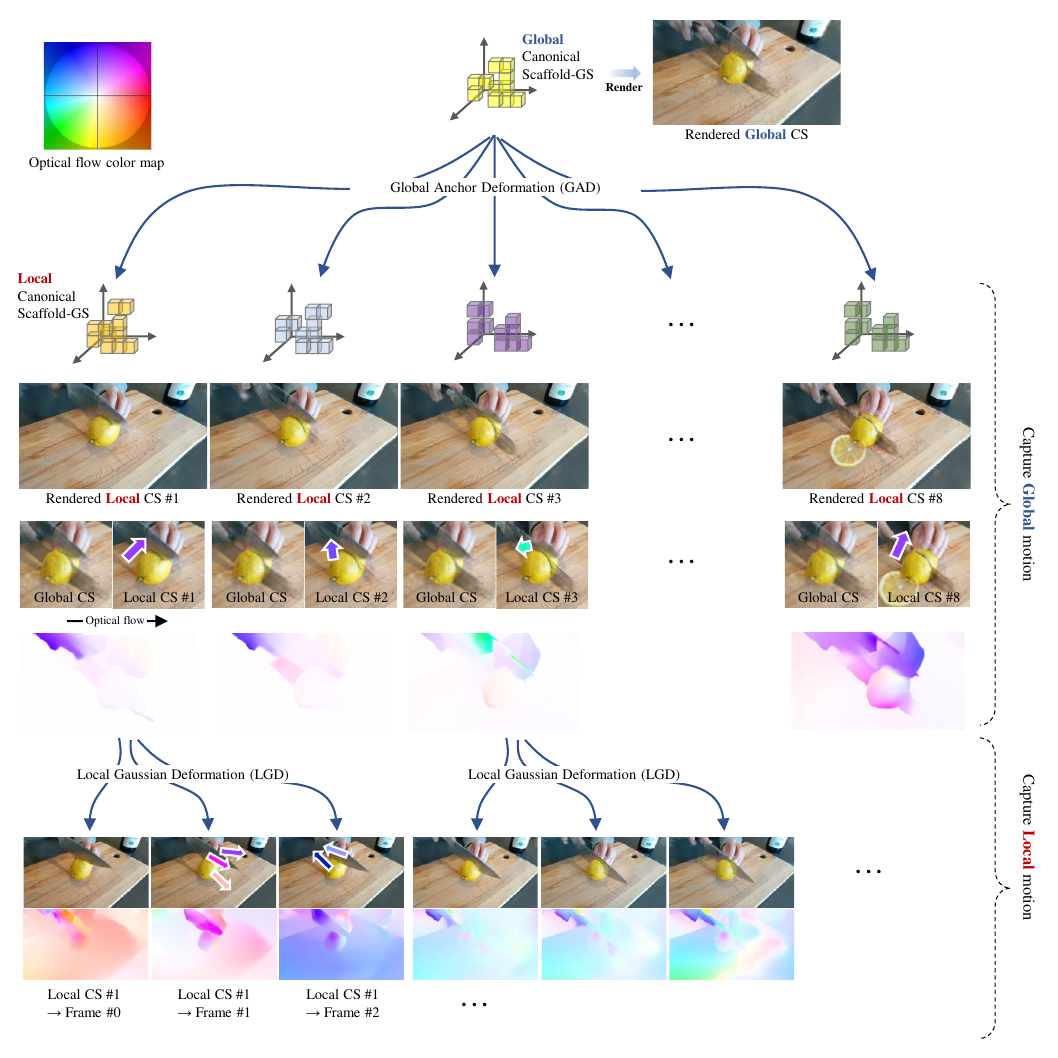}
\vspace{-0.3cm}
\caption{\textbf{Visualization of GLMD.} For \textit{cut-lemon} scene in HyperNeRF \cite{Park2021HyperNeRF} dataset, the rendered patch of Global CS, Local CS, and each time stamp are presented for a fixed camera viewpoint. We also illustrate the optical flow color map between those patches to observe the captured motion at each deformation stage. At GAD stage, deformation in mainly found near objects with dominant motion (e.g., the lemon and knife), and the overall color trends are similar, indicating a similar global motion direction. In contrast, at the LGD stage, motion is observed across the entire scene, with relatively more diverse range of motion directions.}
\label{fig:sub_fig1}
\vspace{-0.5cm}
\end{figure*}

Our MoDec-GS is characterized by its ability to decompose global and local motion through a 2-stage deformation process. This technique, called GLMD, enables effective representation of complex motions even with a limited-size hexplane. To verify whether GLMD operates as our design intention, we visualize the individual rendering results of Global CS, Local CS, and the final deformed frame, which is shown in Fig. \ref{fig:sub_fig1}. For the cut-lemon scene in HyperNeRF, we rendered the Global CS directly, as shown in the topmost image. After the Global CS is deformed into each Local CS through GAD, we rendered each Local CS as shown in the central image and then measured the optical flow \cite{shi2023flowformer++} between the two. As we can see in the rendered Local CS and the optical flow, it can be observed that a global motion with an overall similar direction is represented according to the movement of the knife cutting the lemon. Based on the optical flow color map, we visualized this by overlaying arrows on the rendered patch. The Local CS is then deformed into individual frames through LGD. We also rendered the frames at a fixed camera position during this process and observed the optical flow. As a result, various directional components of local motion were observed, which were also overlaid as arrows on the rendered patch. Through this detailed and intuitive visualization, we confirmed that the proposed GLMD effectively captures both global and local motions. Thanks to this capability, it achieves high scene representation for complex motions even with a smaller model size. 

\subsection{Analysis of Complex Motion through GLMD}
\label{Analysis of Complex Motion through GLMD}
In our work, global motion refers to \textit{rigid transformations} within a time interval, while local motion captures \textit{non-rigid deformations} between consecutive time steps. Complex motion is defined as a combination of these two types of motion. These characteristics can be observed in Fig. \ref{sup:visualization of GLMD}. To further investigate this, we measured the average normalized optical flow magnitudes within an object mask \cite{ravi2024sam} for global motion modeled by GAD, and local motion modeled by LGD. The results are shown in Fig . \ref{fig:Optical flow analysis for GLMD}. As seen in the figure, GAD is primarily associated with object-centric rigid transformations, exhibiting a higher optical flow magnitude in the object mask regions on average. In contrast, LGD distributes the optical flow magnitude across the entire scene with relatively smaller values.

\section{Limitations and Future Works}
\label{sub:Limitations}

\begin{figure*}[t!]
\centering
\includegraphics[scale=0.6]{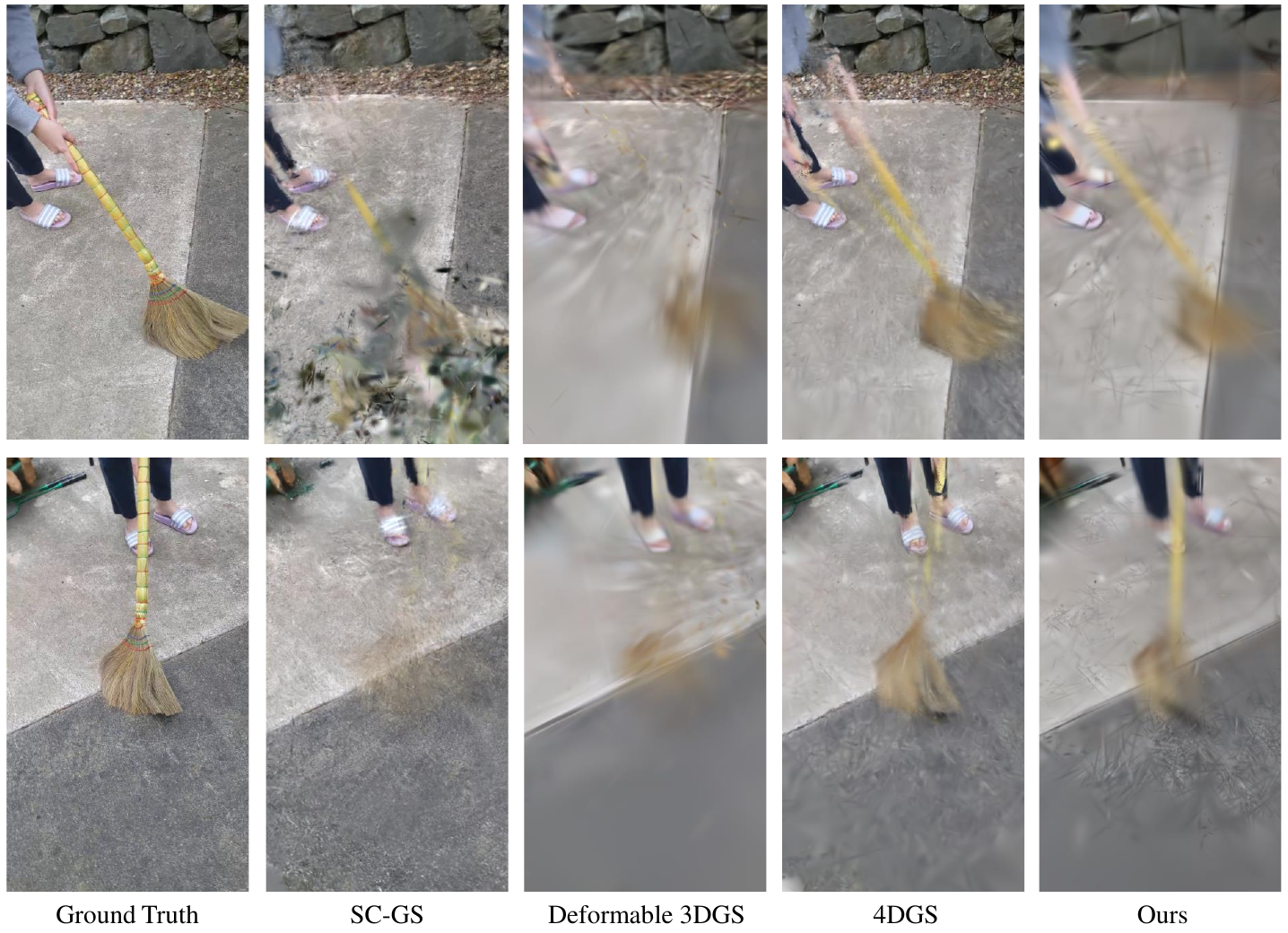}

\caption{\textbf{Failure case: HyperNeRF-broom.} In the face of challenges in reconstructing dynamic scenes from monocular video, there are limitations in adequately representing thin and highly intricate textured objects.}
\label{fig:sub_fig2}

\end{figure*}

Fig. \ref{fig:sub_fig2} illustrates a failure case on the HyperNeRF-broom dataset. In the challenging context of monocular video, representing thin and highly detailed textured objects using a finite number of 3D Gaussians remains a limitation. Consequently, neither the comparison methods \cite{Huang2024SCGS, Yang2024Deformable3DGS, Wu20244DGS} nor ours are able to effectively learn the scene. 

To address this issue, previous studies have explored integrating traditional graphics techniques such as texture and alpha mapping into 3DGS \cite{chao2024textured}, utilizing generalized exponential functions instead of 3D Gaussians \cite{hamdi2024ges}, or incorporating hierarchical pyramid features to enhance detail representation \cite{franketrips2024trips}. As part of future work, we aim to enhance the Gaussian primitives used in MoDec-GS by building upon these prior studies, enabling robust expressivity even in scenes with intricate and highly detailed textures.

\begin{table*}[!h]
\begin{center}
\setlength\tabcolsep{5pt} % default value: 6pt
\renewcommand{\arraystretch}{1.1}
\scalebox{0.68}{
\begin{tabular}{ c  |cc cc cc cc}
\bottomrule \hline
& \multicolumn{8}{c}{(a) iPhone dataset} \\

\noalign{\smallskip}
Method & \multicolumn{2}{c}{Apple}& \multicolumn{2}{c}{Block}& \multicolumn{2}{c}{Paper-windmill}& \multicolumn{2}{c}{Space-out}\\  
\hline\noalign{\smallskip}

SC-GS~\cite{Huang2024SCGS}    
& 14.96 / 0.692 / 0.508 / 0.704 & 173.3 
& 13.98 / 0.548 / 0.483 / 0.931 & 115.7
& 14.87 / \best{0.221} / 0.432 / 0.473 & 446.3
& \best{14.79} / 0.511 / \best{0.440} / 0.411 & 114.2\\

Deformable 3DGS ~\cite{Yang2024Deformable3DGS}       
& \second{15.61} / \second{0.696} / \best{0.367} / \second{0.523} & 87.71 
& \second{14.87} / \second{0.559} / \best{0.390} / \second{0.924} & 118.9
& \second{14.89} / 0.213 / \best{0.341} /0.519 & 160.2
& 14.59 / 0.510 / \second{0.450} / 0.562 & \second{42.01}\\

4DGS~\cite{Wu20244DGS}   
& 15.41 / 0.691 / 0.524 / 0.591 & \second{61.52}
& 13.89 / 0.550 / 0.539 / 1.095& \second{63.52}
& 14.44 / 0.201 / 0.445 / \second{0.375} & 123.9
& 14.29 / \second{0.515} / 0.473 / \second{0.331} & 52.02\\

\textbf{MoDec-GS (Ours)}
& \best{16.48} / \best{0.699} / \second{0.402} / \best{0.459} & \best{23.78} 
& \best{15.57} / \best{0.590} / \second{0.478} / \best{0.852} & \best{13.65} 
& \best{14.92} / \second{0.220} / \second{0.377} / \best{0.357} & \best{17.08} 
& \second{14.65} / \best{0.522} / 0.467 / \best{0.310} & \best{18.24}
\end{tabular}
}

\scalebox{0.68}{
\begin{tabular}{ c  |cc cc cc |cc}
\hline\noalign{\smallskip}
 & \multicolumn{2}{c}{Spin}& \multicolumn{2}{c}{Teddy}& \multicolumn{2}{c}{Wheel}& \multicolumn{2}{c}{\textbf{Average}}\\  
\hline\noalign{\smallskip}

SC-GS~\cite{Huang2024SCGS}    
& 14.32 / 0.407 / 0.445 / \best{1.191} & 219.1
& \second{12.51} / \second{0.516} / \best{0.562} / \second{1.095} & 318.7
& \second{11.90} / \second{0.354} / 0.484 / \second{1.623} & 239.2
& \second{13.90} / \second{0.464} / 0.479 / \second{0.923} & 232.4\\

Deformable 3DGS ~\cite{Yang2024Deformable3DGS}    
& 13.10 / 0.392 / 0.490 / 1.482 & 133.9
& 11.20 / 0.508 / \second{0.573} / 1.460 & 117.1
& 11.79 / 0.345 / \best{0.394} / 1.732 & 106.1
& 13.72 / 0.461 / \best{0.430} / 1.029 & 109.4\\

4DGS~\cite{Wu20244DGS}  
& \second{14.89} / \second{0.413} / \second{0.441} / 1.362 & \second{71.80} 
& 12.31 / 0.509 / 0.605 / 1.156 & \second{80.44} 
& 10.83 / 0.339 / 0.538 / 2.007 & \second{96.50} 
& 13.72 / 0.460 / 0.509 / 0.988 & \second{78.54}\\

\textbf{MoDec-GS (Ours)}  
& \best{15.53} / \best{0.433} / \best{0.366} / \second{1.265} & \best{26.84} 
& \best{12.56} / \best{0.521} / 0.598 / \best{1.056} & \best{12.28} 
& \best{12.44} / \best{0.374} / \second{0.413} / \best{1.561} & \best{16.68} 
& \best{14.60} / \best{0.480} / \second{0.443} / \best{0.837} & \best{18.37}\\

\bottomrule \hline

\end{tabular}
}

%---------------------------------- hypernerf- interp
\scalebox{0.68}{
\begin{tabular}{ c  |cc cc cc cc}
& \multicolumn{8}{c}{(b) Hypernerf dataset} \\
 & \multicolumn{2}{c}{\textit{interp} - Aleks-teapot}& \multicolumn{2}{c}{ \textit{interp} - Chickchiken}& \multicolumn{2}{c}{\textit{interp} - Cut-lemon}& \multicolumn{2}{c}{\textit{interp} - Hand}\\  
\hline\noalign{\smallskip}

SC-GS~\cite{Huang2024SCGS} & 
24.86 / 0.854 / \second{0.186} / 5.406 & 426.0 &
26.05 / 0.781 / \best{0.239} / \best{4.176} & 101.2 &
29.63 / \second{0.862} / \second{0.182} / \second{2.469} & 130.8 &
28.97 / \second{0.859} / 0.192 / \best{4.206} & 404.3 \\

Deformable 3DGS ~\cite{Yang2024Deformable3DGS}  &
20.13 / 0.625 / 0.479 / 11.00 & 108.0 &
25.89 / 0.782 / 0.272 / \second{4.539} & 50.77 &
28.61 / 0.792 / 0.269 / 3.936 & 82.65 &
28.91 / 0.855 / \second{0.191} / 4.574 & 144.6 \\

4DGS~\cite{Wu20244DGS}  &
\best{26.99} / \second{0.853} / 0.193 / \second{3.309} & \second{105.6} &
\best{26.88} / \best{0.797} / 0.336 / 7.036 & \second{50.34} &
\second{30.17} / 0.776 / 0.325 / 5.598 & \second{56.05} &
\best{29.87} / 0.847 / 0.223 / 4.928 & \second{85.26} \\

\textbf{MoDec-GS (Ours)}  & 
\second{26.72} / \best{0.871} / \best{0.162} / \best{3.074} & \best{55.69} &
\second{26.65} / \second{0.793} / \second{0.271} / 4.884 & \best{31.17} &
\best{31.08} / \best{0.878} / \best{0.161} / \best{2.462} & \best{25.40} &
\second{29.65} / \best{0.867} / \best{0.187} / \second{4.355} & \best{73.60} \\

\end{tabular}
}

\scalebox{0.68}{
\begin{tabular}{ c  |cc cc| cc cc}
\hline\noalign{\smallskip}
 & \multicolumn{2}{c}{\textit{interp} - Slice-banana} 
 & \multicolumn{2}{c}{\textit{interp} - Torchocolate} 
 & \multicolumn{2}{c}{\textit{misc} - Americano} 
 & \multicolumn{2}{c}{\textit{misc} - Cross-hands} \\  
\hline\noalign{\smallskip}

SC-GS~\cite{Huang2024SCGS} &
24.57 / 0.641 / \best{0.323} / \best{7.697} & 76.15 & 
\second{27.62} / \second{0.893} / \second{0.155} / \best{2.640} & 217.0 &
30.84 / 0.928 / 0.101 / 3.055 & 271.4 &
\best{28.78} / \best{0.844} / \best{0.198} / \best{2.209} & 222.1 \\

Deformable 3DGS ~\cite{Yang2024Deformable3DGS}  & 
\second{24.74} / 0.647 / \second{0.380} / \second{8.594} & 52.10 & 
27.47 / 0.890 / 0.171 / 2.924 & \second{84.52} & 
\second{30.87} / \second{0.929} / \best{0.094} / \best{2.896} & 141.6 &
27.70 / 0.813 / \second{0.246} / \second{2.683} & 142.8 \\

4DGS~\cite{Wu20244DGS}  & 
\best{25.27} / \best{0.676} / 0.428 / 11.10 & \second{47.45} &
25.44 / 0.829 / 0.301 / 6.784 & 91.10 &
\best{31.30} / 0.917 / 0.137 / 3.706 & \second{85.72} &
28.06 / 0.763 / 0.350 / 6.644 & \second{62.10} \\

\textbf{MoDec-GS (Ours)}  & 
24.70 / \second{0.653} / 0.428 / 8.729 & \best{31.74} &
\best{27.86} / \best{0.896} / \best{0.136} / \second{2.657} & \best{27.34} &
30.55 / \best{0.932} / \second{0.100} / \second{2.934} & \best{43.99} &
\second{28.39} / \second{0.821} / 0.253 / 4.545 & \best{23.97} \\

\end{tabular}
}

\scalebox{0.68}{
\begin{tabular}{ c  |cc cc cc cc}
\hline\noalign{\smallskip}
 & \multicolumn{2}{c}{\textit{misc} - Espresso}
 & \multicolumn{2}{c}{\textit{misc} - Keyboard} 
 & \multicolumn{2}{c}{\textit{misc} - Oven-mitts} 
 & \multicolumn{2}{c}{\textit{misc} - Split-cookie} \\  
\hline\noalign{\smallskip}

SC-GS~\cite{Huang2024SCGS} &
\best{26.52} / \best{0.910} / \best{0.167} / \best{5.162} & 160.4 &
28.47 / \second{0.904} / \best{0.129} / \best{3.980} & 229.4 &
27.54 / \second{0.830} / \second{0.182} / \second{3.483} & 88.63 &
\best{33.01} / \best{0.940} / \best{0.087} / 2.529 & 255.1 \\

Deformable 3DGS ~\cite{Yang2024Deformable3DGS}  & 
25.47 / 0.899 / 0.179 / \second{5.513} & 60.93 &
28.15 / 0.900 / 0.137 / \second{4.190} & 97.77 &
27.51 / \best{0.832} / \best{0.175} / \best{3.396} & {39.83} &
32.63 / \second{0.937} / \second{0.087} / 2.417 & 107.9 \\

4DGS~\cite{Wu20244DGS}  & 
25.82 / 0.899 / 0.191 / 5.732 & 72.93 &
\second{28.64} / 0.895 / 0.177 / 4.762 & 62.57 &
\best{27.99} / 0.801 / 0.316 / 6.241 & \second{45.73} &
32.64 / 0.919 / 0.147 / 3.362 & \second{67.00} \\

\textbf{MoDec-GS (Ours)}  & 
\second{26.16} / \second{0.905} / \second{0.170} / 5.808 & \best{25.06} &
\best{28.68} / \best{0.906} / \second{0.136} / 4.230 & \best{25.63} &
\second{27.78} / 0.820 / 0.220 / 4.630 & \best{20.03} &
\second{32.84} / 0.935 / 0.093 / \best{2.400} & \best{45.88} \\

\end{tabular}
}

\scalebox{0.68}{
\begin{tabular}{ c  |cc |cc cc cc}
\hline\noalign{\smallskip}
 & \multicolumn{2}{c}{\textit{misc} - Tamping}
 & \multicolumn{2}{c}{\textit{vrig} - 3dprinter} 
 & \multicolumn{2}{c}{\textit{vrig} - Broom} 
 & \multicolumn{2}{c}{\textit{vrig} - Chicken} \\
 \hline\noalign{\smallskip}

SC-GS~\cite{Huang2024SCGS} &
23.10 / 0.781 / \best{0.326} / \second{6.352} & 259.4 &
18.79 / 0.613 / \second{0.269} / 15.17 & 101.7 &
18.66 / 0.269 / \best{0.505} / 14.12 & 122.6 &
21.85 / 0.616 / \second{0.257} / 11.83 & 111.2 \\

Deformable 3DGS ~\cite{Yang2024Deformable3DGS}  & 
23.95 / \second{0.804} / \second{0.331} / 6.409 & \best{17.92} &
20.33 / 0.666 / 0.306 / \second{14.11} & \second{40.33} &
21.00 / \second{0.306} / 0.646 / \second{13.12} & 181.8 &
22.66 / 0.642 / 0.276 / 11.12 & 63.25 \\

4DGS~\cite{Wu20244DGS}  & 
\second{24.15} / 0.801 / 0.342 / 6.656 & 78.26 &
\second{21.97} / \second{0.704} / 0.328 / 14.92 & 55.82 &
\best{21.85} / \best{0.365} / \second{0.559} / \best{9.279} & \second{51.13} &
\second{28.53} / \second{0.807} / 0.295 / \second{8.137} & \second{46.11} \\

\textbf{MoDec-GS (Ours)}  & 
\best{24.33} / \best{0.809} / 0.339 / \best{6.329} & \second{24.77} &
\best{22.00} / \best{0.706} / \best{0.265} / \best{13.06} & \best{26.60} &
\second{21.04} / 0.303 / 0.666 / 13.50 & \best{30.83} &
\best{28.77} / \best{0.834} / \best{0.197} / \best{4.936} & \best{23.22} \\

\end{tabular}
}

\scalebox{0.68}{
\begin{tabular}{ c  |cc |cc cc cc}
\hline\noalign{\smallskip}
 & \multicolumn{2}{c}{\textit{vrig} - Peel-banana} 
 & \multicolumn{2}{c}{\textbf{Average - \textit{interp}}} 
 & \multicolumn{2}{c}{\textbf{Average - \textit{misc}}} 
 & \multicolumn{2}{c}{\textbf{Average - \textit{vrig}}} \\
 \hline\noalign{\smallskip}

SC-GS~\cite{Huang2024SCGS} &
25.49 / 0.806 / 0.215 / 4.568 & 519.9 &
26.95 / \second{0.815} / \best{0.213} / \second{4.432} & 226.0 &
28.32 / \best{0.876} / \best{0.170} / 3.824 & 212.3 &
21.19 / 0.575 / \best{0.311} / 11.42 & 231.9 \\

Deformable 3DGS ~\cite{Yang2024Deformable3DGS}  & 
26.93 / \second{0.851} / \second{0.193} / 4.386 & 268.0 &
25.96 / 0.766 / 0.294 / 5.929 & 87.13 &
28.04 / 0.873 / \second{0.178} / \second{3.929} & 86.97 &
22.72 / 0.616 / 0.355 / 10.68 & 138.3 \\

4DGS~\cite{Wu20244DGS}  & 
\second{27.66} / 0.847 / 0.206 / \second{4.179} & \second{93.02} &
\second{27.44} / 0.797 / 0.302 / 6.459 & \second{72.65} &
\second{28.37} / 0.857 / 0.237 / 5.301 & \second{67.76} &
\second{25.00} / \best{0.680} / 0.347 / \second{9.131} & \second{61.52} \\

\textbf{MoDec-GS (Ours)}  & 
\best{28.25} / \best{0.873} / \best{0.171} / \best{3.801} & \best{29.80} &
\best{27.78} / \best{0.827} / \second{0.219} / \best{4.360} & \best{40.82} & 
\best{28.39} / \second{0.875} / 0.187 / \best{4.411} & \best{29.90} &
\best{25.01} / \second{0.679} / \second{0.324} / \best{8.827} & \best{27.61} \\

\bottomrule \hline

\end{tabular}
}

\scalebox{0.68}{
\begin{tabular}{ c  |lc |lc |lc |cc}
 & \multicolumn{8}{c}{(c) Nvidia monocular}\\

\noalign{\smallskip}
Method & \multicolumn{2}{c}{Balloon1}& \multicolumn{2}{c}{Balloon2}& \multicolumn{2}{c}{Jumping}& \multicolumn{2}{c}{dynamicFace}\\
\hline\noalign{\smallskip}
4DGS~\cite{Wu20244DGS}   
& 25.46 / 0.856 / 0.198 / \textcolor{white}{0} - \, \textcolor{white}{0} & 67.43 
& 27.12 / 0.842 / 0.151 / \textcolor{white}{0} - \, \textcolor{white}{0} & 58.36 
& 22.43 / 0.842 / 0.264 / \textcolor{white}{0} - \, \textcolor{white}{0} & 46.19
&  27.32 / 0.935 / 0.121/ \textcolor{white}{0} - \, \textcolor{white}{0} & 123.8
 \\

\textbf{\,\,\,\, MoDec-GS (Ours) \,\,\,\,}            
& \best{26.35} / \best{0.884} / \best{0.173} / \textcolor{white}{0} - \, \textcolor{white}{0} & \best{38.67}
& \best{27.18} / \best{0.875} / \best{0.101} / \textcolor{white}{0} - \, \textcolor{white}{0} & \best{41.37}
&  \best{23.14} / \best{0.858} / \best{0.226} / \textcolor{white}{0} - \, \textcolor{white}{0}  & \best{29.09}
&  \best{29.65} / \best{0.955} / \best{0.094} / \textcolor{white}{0} - \, \textcolor{white}{0} & \best{46.57}
\\
\end{tabular}
}
\scalebox{0.68}{
\begin{tabular}{ c  |lc |lc |lc |cc}
\hline\noalign{\smallskip}
%\noalign{\smallskip}
 & \multicolumn{2}{c}{Playground}& \multicolumn{2}{c}{Skating}& \multicolumn{2}{c}{Truck}& \multicolumn{2}{c}{Umbrella}\\
\hline
4DGS~\cite{Wu20244DGS}   
&  22.17 / 0.743 / 0.215 / \textcolor{white}{0} - \, \textcolor{white}{0} & 81.94
&  28.94 / 0.932 / 0.195 / \textcolor{white}{0} - \, \textcolor{white}{0} & 42.08
&  28.28 / 0.889 / 0.234 / \textcolor{white}{0} - \, \textcolor{white}{0} & 53.69
&  24.80 / 0.714 / 0.297 / \textcolor{white}{0} - \, \textcolor{white}{0}  & 65.96\\

\textbf{\,\,\,\,  MoDec-GS (Ours) \,\,\,\,}            
& \best{23.35} / \best{0.817} / \best{0.149} / \textcolor{white}{0} - \, \textcolor{white}{0}  & \best{49.41} 
& \best{29.31} / \best{0.942} / \best{0.155} / \textcolor{white}{0} - \, \textcolor{white}{0} 
& \best{25.27}
& \best{29.21} / \best{0.911} / \best{0.184} / \textcolor{white}{0} - \, \textcolor{white}{0}  & \best{37.68}
& \best{25.04} / \best{0.762} / \best{0.223} / \textcolor{white}{0} - \, \textcolor{white}{0}  & \best{49.08}\\

\bottomrule \hline
\end{tabular}
}

\caption{\textbf{Quantitative results comparison on (a) iPhone \cite{Gao2022Dycheck}, (b) HyperNeRF \cite{Park2021HyperNeRF}, (c) Nvidia \cite{yoon2020novel} datasets}. \textcolor{red}{\textbf{Red}} and \textcolor{blue}{\underline{blue}} denote the best and second best performances, respectively. Each block element of 5-performance denotes (PSNR(dB)$\uparrow$ / SSIM$\uparrow$ \cite{wang2004image} / LPIPS$\downarrow$ \cite{zhang2018unreasonable} / tOF$\downarrow$ \cite{chu2020learning} \,  Storage(MB)$\downarrow$). For iPhone dataset, the masked metrics are used. For Nvidia monocular dataset, tOF values are not computed since the test views are sparsely distributed along the temporal axis.} 

\label{table:sub_all_results}
\end{center}
\end{table*}

